\documentclass[lettersize, journal]{IEEEtran}
\usepackage{amsmath,amssymb,amsfonts,bm}
\usepackage{algorithm,algorithmic}
\usepackage{array}
\usepackage{booktabs}
\usepackage{multirow}
\usepackage{tabularx}
\usepackage{threeparttable}
\usepackage{tabularray}
\usepackage{makecell}
\usepackage{adjustbox}

\usepackage{graphicx}
\usepackage[caption=false,font=normalsize,labelfont=sf,textfont=sf]{subfig}
\usepackage{stfloats}

\usepackage{cite}
\usepackage{url}
\usepackage{hyperref}
\hypersetup{
    citecolor={green!60!black}, 
    linkcolor={red!80!black},   
    urlcolor=black
}

\usepackage[T1]{fontenc}
\usepackage{xcolor}
\usepackage[table]{xcolor}

\definecolor{deepred}{RGB}{183, 48, 28} 
\definecolor{deepblue}{RGB}{0, 112, 192}

\usepackage{textcomp}   
\usepackage{verbatim}   
\usepackage{pifont}     

\usepackage{balance}
\usepackage{bbding}
\newcommand{\cmark}{\Checkmark}

\newcommand{\RN}[1]{\MakeUppercase{\romannumeral #1}}

\usepackage{enumitem}
\usepackage{lineno}

\def\BibTeX{{\rm B\kern-.05em{\sc i\kern-.025em b}\kern-.08em
    T\kern-.1667em\lower.7ex\hbox{E}\kern-.125emX}}

\begin{document}

\title{HieraRS: A Hierarchical Segmentation Paradigm for Remote Sensing Enabling Multi-Granularity Interpretation and Cross-Domain Transfer}

\author{
  Tianlong~Ai, 
  Tianzhu~Liu,~\IEEEmembership{Member,~IEEE},
  Haochen~Jiang, 
  and~Yanfeng~Gu,~\IEEEmembership{Senior Member,~IEEE}

}

\maketitle

\begin{abstract}
Hierarchical land cover and land use (LCLU) classification aims to assign pixel-wise labels with multiple levels of semantic granularity to remote sensing (RS) imagery. However, existing deep learning-based methods face two major challenges: 1) They predominantly adopt a flat pixel-wise classification paradigm, which limits their ability to generate end-to-end multi-granularity hierarchical predictions aligned with realistic tree-structured hierarchies. 2) Most cross-domain studies focus on performance degradation caused by sensor or scene variations, with limited attention to transferring LCLU models to cross-domain tasks with heterogeneous categories and hierarchies. These limitations hinder the flexibility and generalization of LCLU models in practical applications. To address these challenges, we propose HieraRS, a novel hierarchical segmentation paradigm that enables multi-granularity predictions and supports the efficient transfer of LCLU models to cross-domain segmentation tasks with heterogeneous label hierarchies. We introduce the Bidirectional Hierarchical Consistency Constraint Mechanism (BHCCM), which can be seamlessly integrated into mainstream flat semantic segmentation models to generate hierarchical predictions, while improving both semantic consistency and pixel-wise prediction accuracy. Furthermore, we present TransLU, a dual-branch cross-domain transfer framework comprising two key components: Cross-Domain Knowledge Sharing (CDKS) and Cross-Domain Semantic Alignment (CDSA). TransLU enables dynamic category expansion and facilitates the effective adaptation of LCLU models to heterogeneous hierarchies. In addition, we construct MM-5B, a large-scale multi-source and multi-resolution hierarchical LCLU dataset featuring pixel-wise annotations. Extensive experiments on MM-5B, Crop10m, and WHDLD validate the effectiveness and adaptability of the proposed HieraRS across diverse scenarios. The code and MM-5B dataset will be released at: https://github.com/AI-Tianlong/HieraRS.

\end{abstract}


\section{Introduction}
\label{section:Introduction}
\begin{figure}[!tb]
    \centering 
    \centerline{\includegraphics[width=1.0\linewidth]{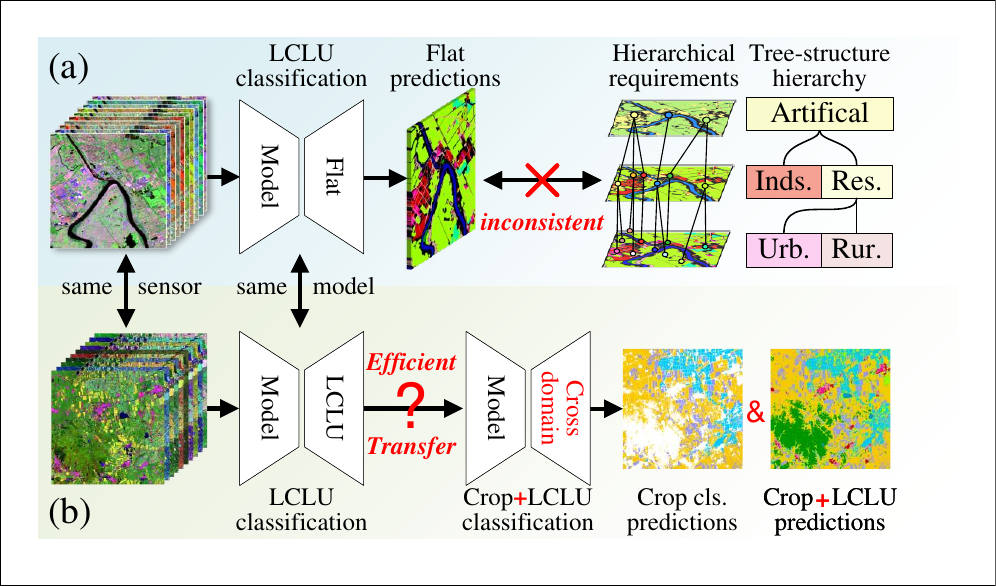}}
    \caption{Illustration of two challenges. (a) Semantic inconsistency between flat classification paradigms and multi-granularity hierarchical requirements. (b) Limited generalization of models with fixed hierarchies when facing dynamic category expansion and cross-domain transfer.}
    \label{fig-challenges}
\end{figure}

\IEEEPARstart{W}{ith} the rapid advancement of aerospace and Earth Observation (EO) technologies, a series of remote sensing (RS) satellites equipped with various sensors, such as Sentinel and GaoFen, have been launched in succession. These satellites collect vast amounts of multi-source and heterogeneous RS data on a daily basis, including multispectral imagery (MSI), hyperspectral imagery (HSI), and synthetic aperture radar (SAR). Characterized by high temporal resolution and global coverage, these data have significantly promoted the widespread adoption of RS in fields such as agricultural monitoring \cite{agricultural_review_2023_NSE}, environmental surveillance \cite{RS_DL_review_3}, and land cover and land use (LCLU) mapping \cite{5B_2023_isprs}. In particular, achieving large-scale LCLU classification with high timeliness and reliability based on RS data is of critical importance for natural resource management \cite{GLC_FCS30D_2023} and urban planning \cite{Cross-city_2023_RSE}.

Deep learning (DL) has played a pivotal role in large-scale EO applications, substantially enhancing the automation of RS data interpretation. In recent years, driven by sustained contributions from the RS community to the development of LCLU datasets, numerous studies have employed DL methods to perform large-scale LCLU classification \cite{SoftFormer_2024_isprs,landcover_TGRS_2022,CFNet_2022, 5B_2023_isprs, Zhang_2019_RSE}. However, as illustrated in Fig.~\ref{fig-challenges}, current DL-based approaches still face two key challenges in practical applications: 

1) \textbf{Deficiency in hierarchical semantic representation and multi-granularity interpretation.} Current approaches commonly treat LCLU classification as a semantic segmentation task, adopting a flat classification paradigm that considers all categories as independent and mutually exclusive, thereby overlooking the inherent hierarchical structure. However, in practical applications, governmental and industrial organizations often construct tree-structured classification systems tailored to specific application needs. For example, the Chinese Ministry of Natural Resources has developed a two-level land use classification system (12/73 categories); the European Union’s CORINE employs a three-level structure (5/15/44 categories); and the UK Environment Agency implements a four-level hierarchy (15/78/150/600 categories). Flat classification methods struggle to produce end-to-end hierarchical predictions that are both semantically consistent and aligned with the target tree-structured label hierarchy, making them difficult to adapt to the aforementioned multi-level interpretation requirements. Furthermore, different application scenarios demand varying levels of interpretation granularity. For instance, natural resource monitoring may only require coarse-grained identification of artificial surfaces, whereas urban-rural land classification necessitates fine-grained subdivision of artificial surfaces into residential, industrial, and transportation categories, which calls for models with multi-granularity hierarchical interpretation capabilities.

2) \textbf{Classification system differences hinder cross-domain generalization.} In addition to the differences in general LCLU classification systems, various domains have also developed dedicated classification standards based on their specific needs. For example, the National Forestry and Grassland Administration of China employs a two-level hierarchy for forest land (8 primary categories and 13 subcategories), while the Ministry of Housing and Urban-Rural Development has established a three-level system for urban land use (8/35/44 categories). Due to discrepancies in category definitions, semantic granularity, and hierarchical structures, existing LCLU models often encounter significant classification biases when directly transferred to tasks in other domains. Furthermore, most existing models cannot incrementally refine a category into finer-grained subcategories without retraining from scratch. For instance, in crop classification tasks, cropland needs to be further subdivided by crop type. However, existing approaches typically incur substantial retraining costs and depend on large amounts of annotated data to support such fine-grained category expansion. In practice, domain-specific applications often suffer from limited labeled samples, which further restricts the generalizability and real-world applicability of these models.

To achieve hierarchical LCLU predictions, several studies have attempted to incorporate the hierarchical category information into model design \cite{yang_2020_b, hiera-unet_2024_TGRS, kang_2024_fusion}. However, most of these methods rely on specific model architectures, making them incompatible with various modern network designs and unable to fully leverage high-performing foundation models recently introduced in the RS community (e.g., SkySense \cite{SkySense_2024_cvpr}, RingMo \cite{RingMo_TGRS_2023}, HyperSIGMA \cite{HyperSIGMA_2025_TPAMI}). Additionally, the adaptability of these methods to multi-source RS data, characterized by varying spatial resolutions and diverse spectral configurations, remains unverified, limiting their potential in complex real-world scenarios. Moreover, recent studies have begun to explore domain adaptation in LCLU tasks, aiming to mitigate domain shifts caused by variations in sensors and geographic regions \cite{cross_domain_1, cross_domain_2, cross_domain_3, cross_domain_4, cross_domain_5}. Although these studies have achieved some progress, they still primarily concentrate on the LCLU task itself. There is still a lack of systematic investigation into how existing LCLU models can be effectively transferred to other EO tasks (e.g., crop classification), which hinders their broader applicability and generalization across diverse EO applications.

To address the aforementioned challenges, this paper proposes HieraRS, a general-purpose hierarchical segmentation paradigm for RS-based EO scenarios. It enables semantically consistent, multi-granularity information extraction and facilitates the efficient transfer of LCLU models to cross-domain segmentation tasks, thereby significantly expanding their application potential. Specifically, we introduce the Bidirectional Hierarchical Consistency Constraint Mechanism (BHCCM), which can be seamlessly integrated into mainstream flat segmentation methods to produce multi-level consistent predictions aligned with complex hierarchical systems, empowering models with robust multi-granularity interpretation capabilities.

To further support effective transfer of LCLU models to segmentation tasks with heterogeneous label hierarchies, we present TransLU, a novel cross-domain transfer framework. TransLU establishes semantic correlations between LCLU tasks and target-domain tasks by leveraging a cross-domain tree-structured hierarchy. By integrating Cross-Domain Knowledge Sharing (CDKS) and Cross-Domain Semantic Alignment (CDSA), this framework supports dynamic category expansion and significantly enhances the adaptability of LCLU models across other diverse mono-temporal segmentation scenarios.

Additionally, to comprehensively evaluate the effectiveness of the proposed HieraRS, we construct MM-5B, a large-scale hierarchical LCLU dataset. The MM-5B dataset includes data from three distinct sensor types with varying spectral configurations and spatial resolutions. It provides pixel-level annotations at three nested hierarchical levels with increasing semantic granularity; specifically, each pixel is labeled at all three levels, which contain 4, 9, and 18 classes, respectively. Together, they form a tree-structured taxonomy with 31 category nodes, spanning from coarse land cover types to fine-grained land use classes.

The contributions of this paper are summarized as follows:
\begin{itemize}
    \item[1)] We construct MM-5B, a large-scale multi-source, multi-resolution hierarchical LCLU dataset with pixel-level labels at three nested levels (4/9/18 classes), serving as a benchmark for hierarchical and cross-resolution LCLU classification.
    \item[2)] We introduce BHCCM, a plug-and-play mechanism that can be seamlessly integrated into mainstream flat segmentation models to produce multi-granularity hierarchical interpretation, improving both semantic consistency and pixel-wise accuracy.
    \item[3)] We propose TransLU, a cross-domain transfer framework with CDKS and CDSA, which supports dynamic category expansion and enables efficient transfer of LCLU models to tasks with heterogeneous label hierarchies.
    \item[4)] We design a hierarchical semantic consistency objective ($\mathcal{L}_\text{HSC}$) with path-level regularization, together with a joint score-based path selection (JSPS) inference strategy to enforce strict tree-structured consistency.
\end{itemize}       

The remainder of this paper is organized as follows. Section~\ref{section:Related Work} reviews representative LCLU datasets and related methods. Section~\ref{section:MM-5B Datasets} describes the construction of MM-5B. Section~\ref{section:Methodology} presents the proposed method and its key components. Section~\ref{section:Experiments and Analysis} reports the experimental setup and results. Finally, Section~\ref{section:Conclusion} concludes the paper and discusses future work.

\section{Related Work} 
\label{section:Related Work}
In this section, we first summarize representative benchmark datasets for LCLU classification. We then review recent advances in pixel-level hierarchical LCLU classification. Finally, we highlight the key challenges of cross-domain transfer in existing LCLU approaches.

\subsection{Pixel-level Annotated LCLU Classification Datasets}
Pixel-level LCLU classification aims to assign a category label to each pixel in RS imagery. To facilitate automated analysis of large-scale RS data, the community has released numerous benchmark datasets for training machine learning and deep learning models. Table~\ref{table-LCLU-datasets} summarizes 12 representative LCLU datasets spanning diverse spatial resolutions and sensing modalities, including RGB~\cite{LoveDA, WHDLD, DeepGlobe}, MSI~\cite{Chesapeake, 5B_2023_isprs}, HSI~\cite{WHU-OHS}, and multi-modal combinations~\cite{SEN12MS, DFC2020}. These datasets cover both urban and rural environments and commonly annotated categories such as buildings, roads, impervious surfaces, cultivated land, and forest.

However, as observed in Table~\ref{table-LCLU-datasets}, existing LCLU datasets still exhibit several limitations, including limited diversity in spatial resolution, restricted category variety, and coarse annotation granularity. These limitations hinder a comprehensive evaluation of hierarchical and cross-resolution LCLU classification methods under multi-source settings. To address this gap, we build MM-5B by extending the fine-grained annotations of the Five-Billion-Pixels dataset~\cite{5B_2023_isprs} with additional imagery from Sentinel-2 and Google Earth, resulting in a large-scale multi-source and multi-resolution hierarchical LCLU benchmark. Details of MM-5B are presented in Section~\ref{section:MM-5B Datasets}.

\begin{table}[tb]
    \centering
    \caption{Comparison of Pixel-Level Annotated Datasets for Land Cover and Land Use Classification}
    \label{table-LCLU-datasets}
    \fontsize{10}{12}\selectfont 
    \renewcommand\arraystretch{1.1} 
    \resizebox{1.0\linewidth}{!}{ 
    \begin{tabular}{lccc}
        \toprule
        \textbf{Dataset} & \textbf{Classes}  & \textbf{Data Source} & \textbf{Resolution} \\
        \midrule   
        LoveDA\cite{LoveDA} & 7 & RGB & 0.3 m \\
        GAMUS\cite{GAMUS}   & 6 & RGB, nDSM & 0.33 m \\
        DeepGlobe\cite{DeepGlobe}   & 7 & RGB & 0.5 m \\
        Chesapeake\cite{Chesapeake} & 6 & RGB, MSI & 2 m \\
        WHDLD\cite{WHDLD} & 6 & RGB & 2 m \\
        BDCI2020\cite{BDCI2020} & 7	& RGB & 2 m \\
        GID\cite{GID}  & 5\,/\,15 & GaoFen-2 & 4 m \\
        Five-Billion-Pixels\cite{5B_2023_isprs} & 24 & GaoFen-2 & 4 m \\
        SEN12MS\cite{SEN12MS} & 17 & Sentinel-1/2 & 10 m \\
        DFC2020\cite{DFC2020} & 10 & Sentinel-1/2 & 10 m \\
        WHU-OHS\cite{WHU-OHS} & 24 & HSI & 10 m \\
        OpenSentinelMap\cite{OpenSentinelMap} & 15 & Sentinel-2 & 10 $\sim$ 60 m \\
        \midrule
        \textbf{MM-5B (ours)} & \makecell{4\,/\,9\,/\,18 \\ (31 total)} & \makecell{RGB, GaoFen-2, Sentinel-2} & 1\,/\,4\,/\,10 m \\
        \bottomrule
    \end{tabular}
    }
\end{table}

\subsection{Hierarchical Land Cover and Land Use Classification}
In remote sensing, DL-based LCLU classification is predominantly formulated as a semantic segmentation task~\cite{land_cover_review_1, land_cover_review_2, land_cover_2, softformer, Crossearth, 5B_2023_isprs, 5B_S2_2023_ICCV, SinoLC-1}. The prevailing paradigm typically adopts a flat taxonomy, treating all classes as mutually exclusive and independent. This assumption, however, disrupts the inherent hierarchical semantic relationships among LCLU types and is often misaligned with the tree-structured classification systems used in operational EO products. While coarse-level predictions can be derived by aggregating fine-level predictions, such a bottom-up strategy is highly sensitive to fine-level classification errors. Errors at lower levels tend to propagate upward through the hierarchy, leading to cumulative deviations in higher-level semantic predictions~\cite{hiera-unet_2024_TGRS}. In addition, without explicit cross-level constraints, flat models may produce semantically inconsistent predictions across levels (e.g., a coarse class that contradicts its fine-level subclasses), further reducing reliability in operational use. Consequently, the flat paradigm fails to satisfy practical hierarchical mapping requirements in terms of cross-level consistency and reliability.

To address the above challenges, recent studies have begun to explore hierarchical LCLU classification methods. Existing efforts can be broadly categorized into hierarchy-aware supervision, architecture-driven hierarchical interaction, and data-driven hierarchy construction. Adam et al.~\cite{Adam_2022_RS} demonstrated through experiments on Sentinel-2 time-series data that hierarchical classification outperforms traditional flat approaches in complex land cover recognition. To more effectively model hierarchical relationships among categories, Yang et al.~\cite{yang_2020_b} introduced class hierarchies derived from geospatial databases, and proposed a joint optimization strategy~\cite{yang_2021_isprs} to enhance hierarchical prediction consistency. HierU-Net~\cite{hiera-unet_2024_TGRS} introduced coarse-level predictions as soft constraints into fine-level classification to achieve hierarchical information transfer. Kang et al.\cite{kang_2024_fusion} extended the hierarchical segmentation framework~\cite{hieraseg_2024_TPAMI} by integrating attribute-based hierarchical structures and a decision-level fusion strategy, leading to improved classification accuracy. In addition, Liu et al.~\cite{clustering_liu_2025} proposed a data-driven clustering method that automatically constructs hierarchical classification systems, which achieved better performance than manually defined hierarchies.

These hierarchical LCLU methods have provided technical support for producing more fine-grained and semantically rich products. However, in modeling hierarchical structures, many of these approaches~\cite{hiera-unet_2024_TGRS} rely on specific architectural designs, limiting their compatibility with modern flat segmentation architectures in hierarchical classification tasks. Moreover, existing studies generally lack a systematic investigation into their adaptability to multi-source RS data characterized by varying spatial resolutions and sensor configurations. To address these challenges, we propose BHCCM, which can be seamlessly integrated into existing flat classification models to achieve end-to-end hierarchical LCLU classification. Details of BHCCM are presented in Section~\ref{Methodology:BHCCM}.

\subsection{Cross-domain Transfer of LCLU Classification Models}
In recent years, extensive research has been devoted to cross-domain adaptation for LCLU classification in RS, primarily aiming to mitigate domain shifts caused by variations in sensors, datasets, geographic regions, and temporal contexts. However, most existing studies focus on transferring LCLU models across domains under the same (or highly overlapping) label space, and thus remain largely confined to the LCLU task itself~\cite{cross_domain_1, cross_domain_2, cross_domain_3, cross_domain_5}. In practical EO pipelines, pretrained LCLU models are often expected to be reused for downstream EO tasks with heterogeneous category definitions and semantic granularities (e.g., crop-type mapping). This setting introduces not only domain shift but also label-system mismatch, including inconsistent granularity and non-isomorphic hierarchical structures, which remains underexplored in existing LCLU transfer studies. As a result, the reusability of LCLU models across diverse segmentation applications is still limited.

To address this gap, we propose TransLU, a dual-branch cross-domain transfer framework designed to bridge LCLU models with diverse segmentation tasks featuring heterogeneous label hierarchies. TransLU leverages hierarchical category structures to establish semantic correspondences across tasks, enabling model reuse under heterogeneous taxonomies rather than assuming identical label spaces. We further provide a systematic empirical study of cross-domain transfer with TransLU on representative tasks, including crop classification and LCLU classification under varying hierarchical systems. The results demonstrate that TransLU achieves consistent improvements over competitive baselines across these two cross-domain EO scenarios, indicating its effectiveness for cross-task transfer under hierarchical label mismatch. Details of TransLU are presented in Section~\ref{Methodology:TransLU}.

\section{The MM-5B Dataset}
\label{section:MM-5B Datasets}
This section introduces the MM-5B dataset, a large-scale multi-source and multi-resolution hierarchical LCLU dataset.
\subsection{Data Sources and Construction Workflow}
The MM-5B dataset is an extension of the Five-Billion-Pixels dataset~\cite{5B_2023_isprs}, constructed by incorporating additional imagery from Sentinel-2 and Google Earth. It constitutes a multi-source and multi-resolution hierarchical LCLU dataset with varied spatial resolutions and spectral configurations.

As the label foundation of MM-5B, Five-Billion-Pixels provides large-scale, quality-controlled pixel-level annotations. It contains 150 GaoFen-2 images with a total geographical coverage of over 50,000 km², where more than 5 billion pixels are carefully annotated. Its category system includes 24 fine-grained LCLU classes, covering artificially constructed, agricultural, and natural categories; importantly, it also contains several land use subclasses refined from the national standard GB/T 21010-2017 (e.g., stadium, square, road, overpass, station, airport), which enriches practical application scenarios. In addition, the imagery is collected from more than 60 dispersed administrative districts across China, enabling broad geographic diversity in landscapes and environmental conditions. Crucially, the creation of Five-Billion-Pixels relied entirely on manual annotation by domain experts. To maximize label consistency and minimize human errors, the annotation process involved a rigorous four-stage workflow: coarse labeling, fine labeling, fine checking, and spot checking. In the final round of quality assurance, a spot check covering 10\% of the samples was conducted, which revealed no significant errors.

Building upon these annotations, MM-5B is constructed from multi-source satellite imagery collected from Google Earth, GaoFen-2, and Sentinel-2, each with distinct spatial resolutions and spectral configurations. Specifically, Google Earth imagery provides 1-meter resolution with three spectral bands (RGB); GaoFen-2 provides MSI with four bands (RGB and NIR) at 4-meter resolution; and Sentinel-2 imagery delivers 10-band multispectral inputs (B2–B8, B8A, B11, and B12) with a spatial resolution of 10 meters. The spatial and spectral complementarity of these distinct sources effectively supports a wide range of typical application scenarios in optical RS, offering rich information for cross-sensor and cross-resolution model development.

\begin{figure}[t]
    \centering 
    \centerline{\includegraphics[width=0.95\linewidth]{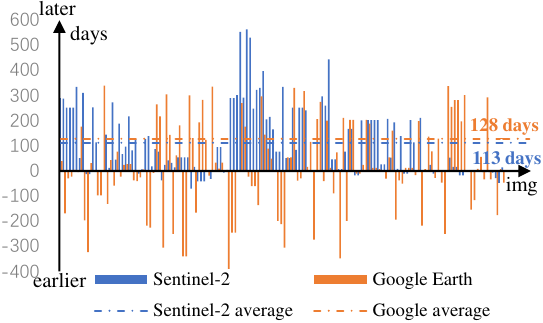}}
    \caption{Distribution of acquisition time differences between the primary GaoFen-2 imagery and the supplementary Sentinel-2 and Google Earth sources in the MM-5B dataset.}
    \label{fig-MM-5B-vs}
\end{figure}

\begin{figure*}[t]
    \centering 
    \centerline{\includegraphics[width=0.95\linewidth]{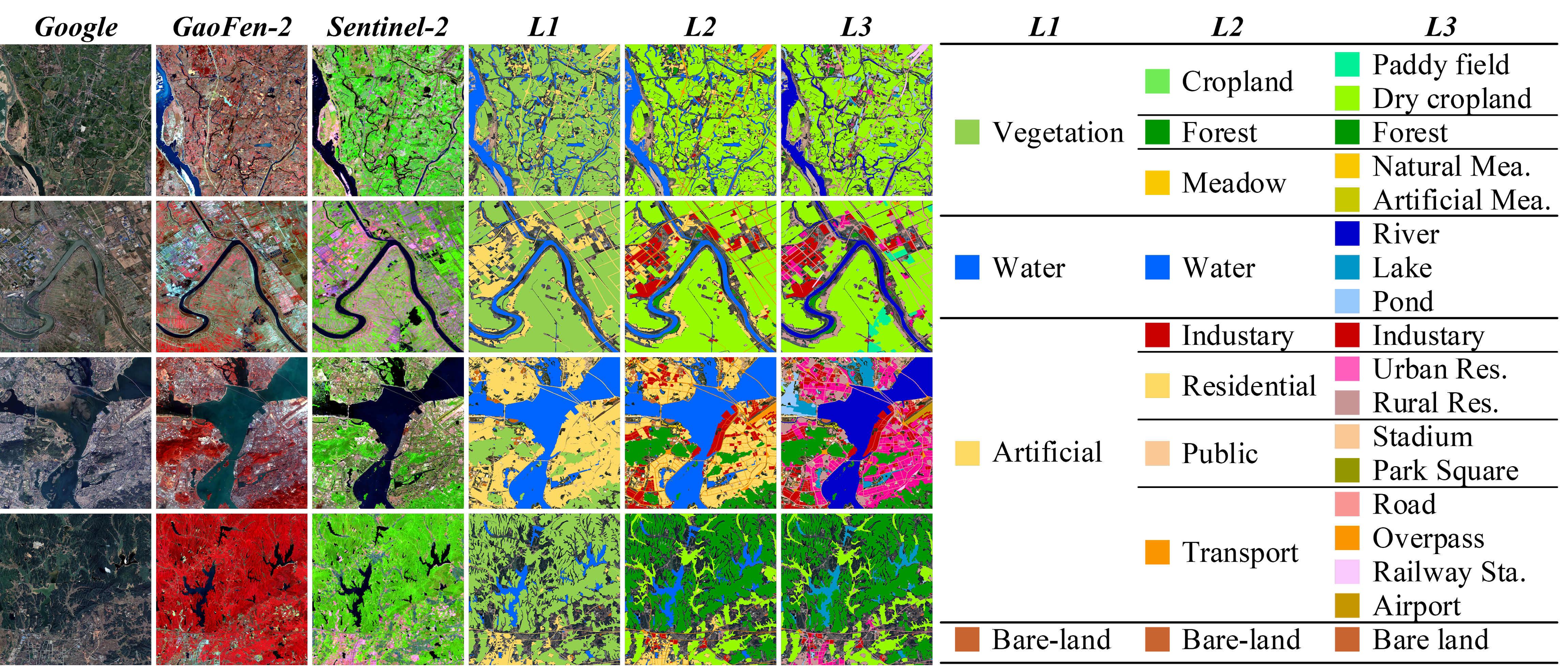}}
    \caption{Illustration of the MM-5B dataset. The left panel shows visualizations of three paired data sources and their corresponding labels at three hierarchical levels: Google Earth (RGB), GaoFen-2 (B4-B3-B2, false color), and Sentinel-2 (B12-B8A-B4, SWIR). The right panel presents the tree-structured hierarchical classification system of the MM-5B, consisting of three levels: $L1$ with 4 categories, $L2$ with 9 categories, and $L3$ with 18 categories.}
    \label{fig-MM-5B-hiera-structure}
\end{figure*}

To ensure accurate spatial alignment across the multi-source images, we designed a two-stage geometric correction and registration pipeline. First, we orthorectified the GaoFen-2 imagery and the corresponding labels using the RPC files provided with the original Five-Billion-Pixels dataset together with the Copernicus 30-m Digital Elevation Model (DEM). Second, we used the Google Earth imagery resampled to 4 m as the reference basemap and co-registered the orthorectified GaoFen-2 images and their labels to achieve fine-grained geometric alignment. All registration results were manually inspected to ensure spatial consistency.

Based on the refined spatial extents of the registered GaoFen-2 scenes, we retrieved the corresponding Google Earth and Sentinel-2 imagery covering the same geographic areas, forming spatially aligned three-source image sets for each scene. To handle temporal disparities among sources, we adopted a prioritized selection strategy: \textit{Image Quality > Temporal Proximity > Seasonal Consistency}. In practice, we prioritize matches within a few months whenever possible. Specifically, since Google Earth imagery is provided as an archive with fixed timestamps, we selected scenes closest to the GaoFen-2 acquisition dates. For Sentinel-2, we retrieved the temporally closest available cloud-free observations. However, because Sentinel-2 was launched later than GaoFen-2, it is difficult to find temporally close Sentinel-2 matches for GaoFen-2 scenes acquired in 2014 and early 2015. In such cases, we substitute Sentinel-2 imagery from the corresponding season in the following year. This strategy prioritizes phenological consistency over absolute temporal proximity. To strictly limit temporal divergence, we excluded combinations where the time difference exceeded two years. Consequently, a total of 136 valid multi-source image triplets were retained. Statistical analysis indicates that the average acquisition time differences between Google Earth and GaoFen-2, and between Sentinel-2 and GaoFen-2, are approximately 128 days and 113 days, respectively, with the full distributions shown in Fig.~\ref{fig-MM-5B-vs}.

Because temporal gaps may introduce land-surface changes, we additionally conducted careful manual inspection across the spatially aligned triplets. For regions where significant LCLU changes were observed between the collected imagery and the GaoFen-2-derived labels, we revised the labels by marking these pixels as "unlabeled", so that they can be excluded during model training and evaluation to reduce the impact of label-image mismatch. This relabeling is applied only when changes are visually evident and cannot be reliably corrected, and it affects only a small fraction of pixels in the dataset.

Finally, the accurately registered 4-m labels were pixel-wise aligned to the 1-m Google Earth and 10-m Sentinel-2 data, forming the multi-source and multi-resolution MM-5B dataset. Fig.~\ref{fig-MM-5B-hiera-structure} presents sample data from MM-5B, highlighting the complementarity of spatial resolution and spectral characteristics across different sources and resolutions.

\subsection{Organization of the Tree-Structured Hierarchy}
One of the key contributions of this paper is the introduction of BHCCM, which can be seamlessly integrated into flat classification methods to enable hierarchical LCLU classification. To support this task, we organize the pixel-level labels of MM-5B into a tree-structured hierarchy that reflects operational EO taxonomies, supporting end-to-end hierarchical LCLU learning and evaluation.

Before organizing the labels into the tree-structured hierarchy, we reorganized the original 24 classes in Five-Billion-Pixels into 18 classes. This adjustment was based on extensive pilot experiments and manual inspections, aiming to assess class separability and annotation usability under different spatial resolutions and data-source conditions. Across multiple baseline experiments on both GaoFen-2 and Sentinel-2, we observed consistently poor performance (mIoU < 30) for several fine-grained categories, and the degradation was more pronounced on Sentinel-2 at 10 m. This indicates that these categories are difficult to distinguish reliably at 10 m resolution. The primary reason is that discriminative cues available at 4 m resolution (e.g., fine-scale textures, boundaries, and object shapes) are substantially weakened after downsampling, which compresses the visual differences between classes. Consequently, retaining these fine-grained categories in a multi-source, multi-resolution setting can lead to unstable learning and reduced cross-source consistency. Based on this evidence, we performed a principled merging of classes that either have insufficient discriminability at 10 m resolution or consistently yield low mIoU in our experiments, resulting in a more robust and interpretable 18-class taxonomy across sources and resolutions.

Specifically, irrigated field and dry cropland were merged into a single cropland class; garden land, shrub forest, and arbor forest were grouped into forest; fish pond and pond were merged into pond; park and square were combined; and the glacier and snow category was removed due to seasonal inconsistencies in acquisition time. As a result, the final MM-5B dataset consists of 136 scenes, each containing spatially aligned multi-source and multi-resolution imagery from three platforms and pixel-wise annotations for 18 LCLU classes.

In MM-5B, the 18 LCLU categories are organized into a three-level hierarchy with increasing semantic granularity from $L1$ to $L3$. Specifically, $L1$ includes four broad classes: vegetation, water, artificial surfaces, and bare land; $L2$ refines them into nine intermediate classes; and $L3$ further divides them into 18 fine-grained categories. The tree-structured hierarchy is illustrated in Fig.~\ref{fig-MM-5B-hiera-structure}. Notably, the three-level hierarchy in MM-5B covers both coarse-grained land-cover categories (e.g., vegetation, water, and artificial surfaces) and fine-grained land-use classes (e.g., airport, urban residential area, and overpass), therefore forming an LCLU taxonomy. During model training, the dataset pipeline dynamically constructs hierarchical labels based on the predefined tree structure, without requiring any manual preprocessing of the hierarchical annotations.

The proposed MM-5B dataset can be used not only for training hierarchical LCLU classification models under single-source settings, but also serves as a novel multi-source and multi-resolution LCLU classification benchmark for the RS community. We believe that leveraging the complementary information from high-resolution optical imagery and medium-resolution MSI can enable more accurate LCLU mapping, holding promising potential for advancing intelligent EO applications.

\section{Methodology}
\label{section:Methodology}

\begin{figure*}[t]
    \centering 
		\centerline{\includegraphics[width=0.95\textwidth]{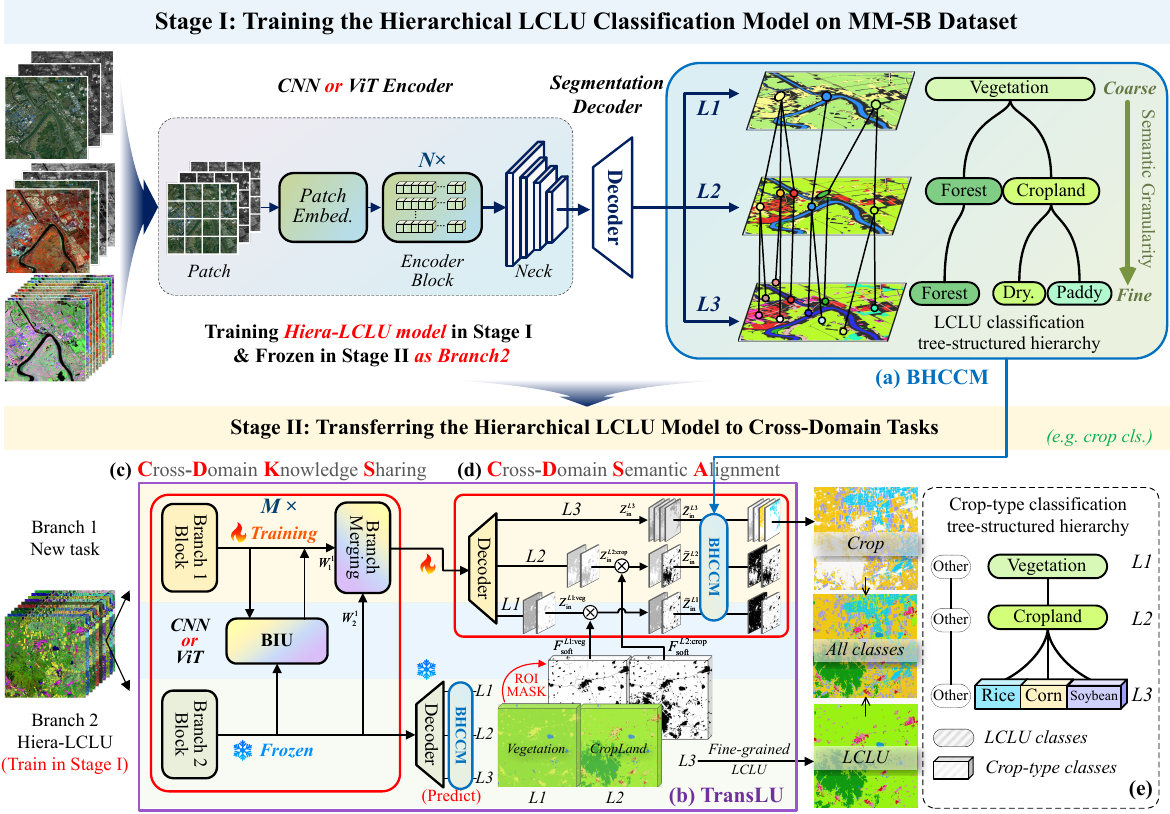}}
		\caption{Overview of HieraRS. Stage~\RN{1}: (a) BHCCM seamlessly integrates into flat semantic segmentation baselines and converts them into hierarchical LCLU classification models. Stage~\RN{2}: (b) TransLU enables cross-domain transfer of the Stage-\RN{1} model under heterogeneous label systems. (c) CDKS performs feature interaction and knowledge sharing across domains; (d) CDSA aligns feature semantics across domains; and (e) shows the cross-domain tree-structured hierarchy used to transfer the hierarchical LCLU model to crop classification.}
    \label{fig-overall_arch}
\end{figure*}

\subsection{Overview of HieraRS}
This section presents HieraRS, a hierarchical segmentation paradigm tailored for RS-based EO applications, designed to facilitate both multi-granularity interpretation and cross-domain task transfer.

Figure~\ref{fig-overall_arch} illustrates the overall architecture of our method. We define HieraRS as a two-stage hierarchical framework consisting of BHCCM and TransLU. HieraRS exploits the intrinsic hierarchical structure of LCLU categories to enforce cross-level prediction consistency and improve transferability to segmentation tasks with heterogeneous label taxonomies. This work pursues two objectives: i) to develop a modular mechanism that can be seamlessly integrated into existing flat semantic segmentation baselines to enable multi-granularity hierarchical LCLU prediction; and ii) to design a generalizable framework that effectively transfers LCLU models across domains under heterogeneous label systems. Accordingly, BHCCM addresses objective i), while TransLU addresses objective ii).

In the first stage, we convert existing flat semantic segmentation baselines into a hierarchical model that produces predictions at multiple label granularities while maintaining cross-level consistency. By training the proposed BHCCM on the MM-5B dataset, we obtain a hierarchical LCLU model capable of multi-granularity interpretation of basic LCLU categories. In the second stage, we transfer the hierarchical model learned in Stage I to new tasks with heterogeneous label taxonomies using the proposed TransLU framework, which consists of two key components, CDKS and CDSA. Stage II in Fig.~\ref{fig-overall_arch}~(b)-(e) illustrates an example of transferring the hierarchical LCLU model to a crop-type classification task.

\subsection{Bidirectional Hierarchical Consistency Constraint Mechanism}
\label{Methodology:BHCCM}

\begin{figure}[t]
    \centering 
    \centerline{\includegraphics[width=0.90\linewidth]{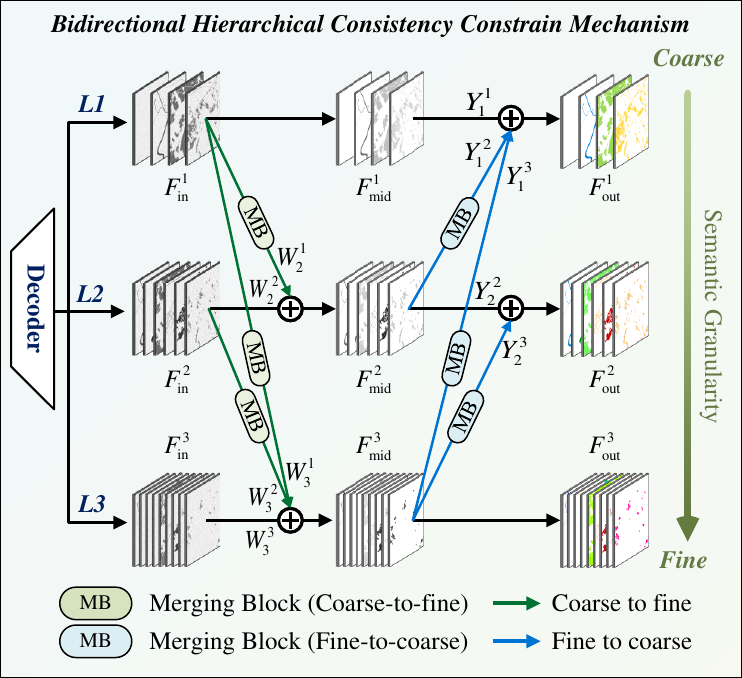}}
    \caption{Detailed structure of BHCCM. The figure illustrates the bidirectional information flow pathways, where semantic feature interaction and fusion across hierarchical levels are achieved via coarse-to-fine (green arrows) and fine-to-coarse (blue arrows) connections.}
    \label{fig-BHCCM}
\end{figure}

The proposed BHCCM is designed to seamlessly integrate into existing flat semantic segmentation architectures, extending them into hierarchical LCLU models capable of producing consistent multi-granularity prediction. As illustrated in Fig.~\ref{fig-overall_arch}(a) and detailed in Fig.~\ref{fig-BHCCM}, BHCCM operates on the output of standard decoders (e.g., UperNet~\cite{upernet_2018_ECCV} and DeepLabv3+~\cite{deeplabv3plus_2018_ECCV}). Unlike conventional approaches, where a simple 2D convolution projects upsampled features from $\mathbb{R}^{B \times C_{\text{dim}} \times H \times W}$ directly to the finest-level space $\mathbb{R}^{B \times C_{\text{L3}} \times H \times W}$, our method replaces this final projection layer with BHCCM to enable parallel prediction at multiple semantic levels. Notably, this represents the only minor modification to the model architecture, while the rest remains unchanged.

Structurally, BHCCM is composed of three key components: i) three level-specific prediction heads that generate initial maps $F^i_{\text{in}}$ ($i = 1, 2, 3$); ii) a coarse-to-fine fusion pathway that injects coarse-grained semantics into fine-grained predictions; and iii) a fine-to-coarse fusion pathway for feeding refined fine-grained semantics back to coarse-grained levels.

Specifically, to initiate the process, BHCCM projects the decoder output through three independent convolutional heads with different output channel sizes, generating initial prediction maps in $\mathbb{R}^{B \times C_{\text{Li}} \times H \times W}$, where the channel dimensions are set to $C_{\text{L1}}=4$, $C_{\text{L2}}=9$, and $C_{\text{L3}}=18$, corresponding to the three hierarchical levels of the MM-5B dataset. These initial prediction features are denoted as $F^i_\text{in}$ and serve as level-specific prediction scores before cross-level fusion.

Following initialization, BHCCM employs bidirectional semantic consistency constraints to enable effective cross-level interaction. In contrast to prior work ~\cite{hieraseg_2024_TPAMI, kang_2024_fusion} that applies hierarchical loss in a passive manner, BHCCM explicitly encodes parent-child semantic dependencies and conducts two-way feature fusion along both coarse-to-fine and fine-to-coarse pathways as depicted by the green and blue arrows in Fig.~\ref{fig-BHCCM}. 

First, in the coarse-to-fine path (indicated by green arrows), coarse-level semantic features are used to guide the refinement of fine-grained predictions. Taking level $L2$ as an example, the intermediate feature $F^2_{\text{mid}}$ consists of two components: i) $F^1_{\text{in}}$ processed through a coarse-to-fine Merging Block (MB) and weighted by the learnable scalars $W^1_2$, and ii) $F^2_{\text{in}}$ weighted by $W^2_2$. Similarly, $F^3_{\text{mid}}$ integrates information from $F^1_{\text{in}}$, $F^2_{\text{in}}$, and $F^3_{\text{in}}$. This process can be formulated as:
\begin{align}
    F^1_{\text{mid}} &= F^1_{\text{in}}  \\
    F^2_{\text{mid}} &= W^1_2 \cdot \text{MB}(F^1_{\text{in}}) + W^2_2 \cdot F^2_{\text{in}}  \\
    F^3_{\text{mid}} &= W^1_3 \cdot \text{MB}(F^1_{\text{in}}) + W^2_3 \cdot \text{MB}(F^2_{\text{in}}) + W^3_3 \cdot F^3_{\text{in}}
\end{align}
where $F^i_{\text{mid}}$ represents the intermediate feature at level $i$, $\text{MB}(\cdot)$ denotes the proposed Merging Block, and $W^j_k$ are learnable fusion weights initialized to 1, controlling the contribution from each level.

Subsequently, in the fine-to-coarse path (indicated by blue arrows), the final output $F^3_{\text{out}}$ is formed by combining the fused fine-level features $F^3_{\text{mid}}$ with the initial features $F^3_{\text{in}}$ weighted by the learnable parameter $Z^3_3$. For the coarser levels $L2$ and $L1$, following the indicated pathways, the final outputs $F^2_{\text{out}}$ and $F^1_{\text{out}}$ are derived by processing the relevant intermediate features through the fine-to-coarse MB and aggregating them with the weighted initial features $F^2_{\text{in}}$ and $F^1_{\text{in}}$. This process can be formulated as:
\begin{align}
    \begin{split} 
        F^1_{\text{out}} &= Y^3_1 \cdot \text{MB}(F^3_{\text{mid}}) + Y^2_1 \cdot \text{MB}(F^2_{\text{mid}}) + Y^1_1 \cdot F^1_{\text{mid}} \\
        &\quad + Z^1_1 \cdot F^1_{\text{in}}
    \end{split} \\
    F^2_{\text{out}} &= Y^3_2 \cdot \text{MB}(F^3_{\text{mid}}) + Y^2_2 \cdot F^2_{\text{mid}} + Z^2_2 \cdot F^2_{\text{in}} \\
    F^3_{\text{out}} &= F^3_{\text{mid}} + Z^3_3 \cdot F^3_{\text{in}}
\end{align}
where $F^i_{\text{out}}$ represents the final hierarchical predictions. The learnable weights $Y^j_k$ and $Z^k_k$ are initialized to 1 and serve to balance the contributions from cross-level fusion and residual connections of input features, respectively.

\begin{figure}[t]
		\centerline{\includegraphics[width=0.90\linewidth]{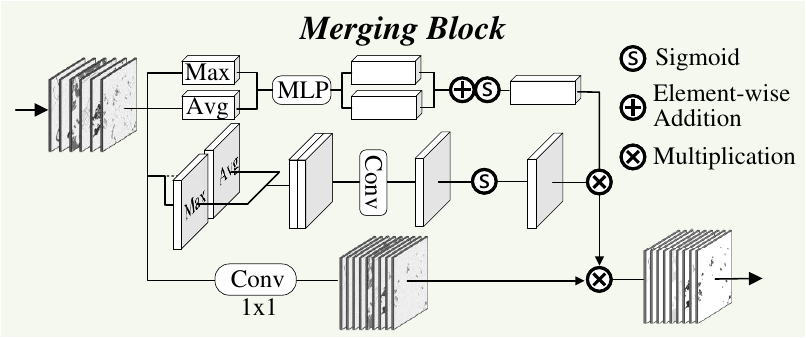}}
		\caption{Detailed structure of the Merging Block. MB consists of three components: a channel-alignment projection, a channel-attention branch, and a spatial-attention branch. The projected features are recalibrated by the channel and spatial attentions to produce dimension-aligned features for cross-level fusion.}
        \label{fig-MB}
\end{figure}

To enable effective information transfer across hierarchical levels, we design a Merging Block (MB) that incorporates both channel and spatial attention to produce intermediate features that are semantically guided and dimensionally aligned with the target level. As illustrated in Fig.~\ref{fig-MB}, MB consists of three components: a channel-alignment projection, a channel-attention branch, and a spatial-attention branch.

\textit{Channel-alignment Projection:} Because features from coarse and fine levels have different channel dimensions, MB first applies a $1\times1$ convolution to project the source features to the target channel dimension. This projection provides an aligned feature space for subsequent attention modulation.

\textit{Spatial Attention Branch:} To explicitly facilitate spatial feature transfer between coarse and fine hierarchical levels, and to convey class-related spatial cues from the current level to the target level, MB computes a spatial attention map to emphasize salient spatial locations. Specifically, we apply channel-wise average pooling and max pooling to the input features, concatenate the resulting two maps, and feed them into a convolution layer followed by a sigmoid activation to obtain the spatial attention map.

\textit{Channel Attention Branch:} To effectively transfer LCLU semantic features from the current level to the target level, the channel attention branch aggregates global context to recalibrate feature channels. Specifically, we apply both global average pooling and global max pooling to the input features to obtain two channel descriptors. These descriptors are passed through a shared MLP, merged via element-wise addition, and projected to the target-level dimensionality, producing a channel attention map that re-weights channels and enhances task-relevant semantics. Finally, the dimensionally aligned features are element-wise multiplied by both the channel and spatial attention maps. This results in a recalibrated feature representation that is strictly aligned with the target level, providing a robust initialization for cross-level fusion. 

In summary, BHCCM serves two roles: in Stage~\RN{1}, it enables multi-granularity hierarchical LCLU prediction by integrating with flat classifiers; in Stage~\RN{2}, it is reused within the CDSA module of TransLU to facilitate cross-domain transfer of LCLU models.

\subsection{Cross-domain Task Transfer Framework---TransLU}
\label{Methodology:TransLU}

We propose TransLU, a novel dual-branch cross-domain transfer framework that enables the hierarchical LCLU model trained in Stage I to be efficiently transferred to cross-domain segmentation tasks with heterogeneous label hierarchies. As illustrated in Fig.~\ref{fig-overall_arch}(b), TransLU consists of two parallel branches. Branch~2 is the pretrained hierarchical LCLU model from the Stage~\RN{1} with frozen parameters during the transfer process. Branch~1 follows the same architecture as Branch~2 but is fine-tuned on the target task with trainable parameters.

To enable knowledge sharing and semantic alignment between the LCLU classification task and the target task, we introduce two key strategies: i) Cross-Domain Knowledge Sharing (CDKS, Fig.~\ref{fig-overall_arch}(c)), which facilitates cross-branch feature interaction between the two encoders; and ii) Cross-Domain Semantic Alignment (CDSA, Fig.~\ref{fig-overall_arch}(d)), which leverages multi-granularity predictions from the frozen LCLU branch to impose semantic constraints on the target branch via a cross-domain tree-structured hierarchy.

\subsubsection{Cross-domain Knowledge Sharing}
\begin{figure}[t]
		\centerline{\includegraphics[width=0.85\linewidth]{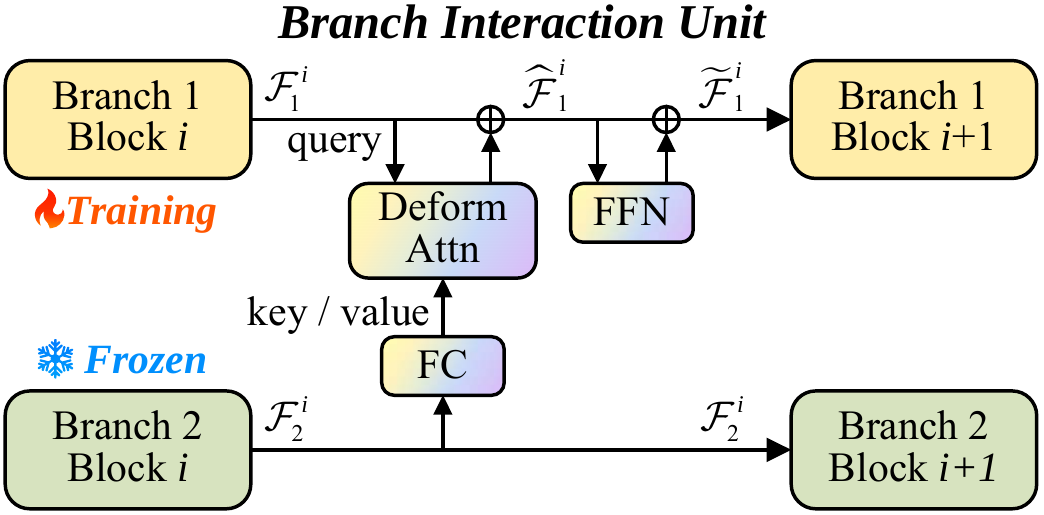}}
		\caption{Detailed structure of the branch interaction unit. It consists of a deformable cross-attention, a fully-connected layer, and a feed-forward network.}
        \label{fig-BIU}
\end{figure}

To promote faster convergence and superior performance in cross-domain transfer, we initialize the encoder of Branch~1 with the pretrained weights of Branch~2 while keeping Branch~2 frozen during training. Inspired by \cite{PIIP_Neurips_2024}, we further propose CDKS to explicitly connect the two branches for effective knowledge sharing. Concretely, CDKS inserts multiple Branch Interaction Units (BIUs) into the encoder, enabling feature interaction and fusion between Branch~1 and Branch~2. The detailed structure of a BIU is presented in Fig.~\ref{fig-BIU}.

Formally, we denote the outputs of the $i$-th encoder block from Branch~1 and Branch~2 as $\mathcal{F}_1^i$ and $\mathcal{F}_2^i$, respectively. To facilitate information interaction, we employ deformable cross-attention \cite{deform_attn_2020}, denoted as $\text{Attn}(\cdot)$ to inject information from Branch~2 into Branch~1. Before the attention operation, a linear projection layer $\text{FC}(\cdot)$ projects the features from Branch~2 into the key and value embedding space used by the cross-attention, facilitating interaction with the query from Branch~1. In practice, $\mathcal{F}_1^i$ serves as the query, and $\text{FC}(\mathcal{F}_2^i)$ provides the key and value for cross-attention. The interaction output is then refined by a feed-forward network $\text{FFN}(\cdot)$. The feature interaction for Branch~1 after the $i$-th block is formulated as:
\begin{align}
    \hat{\mathcal{F}}_1^i &= \mathcal{F}_1^i + \gamma_1^i \text{Attn}(\text{norm}(\mathcal{F}_1^i), \text{norm}(\text{FC}(\mathcal{F}_2^i))), \\
    \widetilde{\mathcal{F}}_1^i &= \hat{\mathcal{F}}_1^i + \tau_1^i \text{FFN}(\text{norm}(\hat{\mathcal{F}}_1^i)).
\end{align}
where $\text{norm}(\cdot)$ denotes LayerNorm~\cite{layernorm_2016}. $\tau_1^i$ and $\gamma_1^i$ are learnable scaling parameters and are initialized to zero to avoid abruptly changing the feature distribution of $\mathcal{F}_1^i$ at the beginning of transfer, thereby better preserving the pretrained knowledge from the LCLU model. This design yields an asymmetric information flow, where Branch~1 continuously absorbs discriminative representations from the frozen Branch~2.

To explicitly distill the robust feature representations from the frozen Branch~2 pretrained on MM-5B into the trainable Branch~1, we introduce a lightweight Branch Merging operation immediately following each feature interaction step. Unlike the deformable cross-attention which injects specific contextual information via residual updates, this merging operation explicitly consolidates the task-adaptive features of Branch~1 with the stable pretrained features of Branch~2, ensuring that the pretrained knowledge is actively propagated to subsequent blocks. Specifically, after obtaining the interacted feature $\widetilde{\mathcal{F}}_1^i$, we form the input to the next block of Branch~1 by adding $\widetilde{\mathcal{F}}_1^i$ and $\mathcal{F}_2^i$ with learnable weights. This fusion acts as a learnable gate that controls the contribution of the frozen branch during transfer, improving training stability and transfer effectiveness.

In practice, the integration frequency adapts to the backbone architecture. For CNN-based models such as ConvNeXt~\cite{convnext_2022_cvpr}, we insert an interaction-and-merging unit after every three convolutional layers, resulting in a total of 12 interaction steps for the ConvNeXt-Base configuration. Conversely, for Transformer-based models such as ViT~\cite{vit_2021}, the operation is performed after each Transformer block. Finally, the outputs from both branches are fed into their respective segmentation decoder for hierarchical prediction, with Branch 2 remaining independent throughout the forward pass.

\subsubsection{Cross-domain Semantic Alignment}
To achieve efficient transfer of hierarchical LCLU classification models to new domain tasks, we propose the CDSA strategy, as illustrated in Fig.~\ref{fig-overall_arch}(d). The core principle of CDSA is to leverage the multi-granularity predictions of the source LCLU model to guide the cross-task semantic alignment. 

First, we construct a cross-domain tree-structured hierarchy based on the categorical relationships between the source LCLU task and the target task. As illustrated in Fig.~\ref{fig-overall_arch} (e), this hierarchy serves as a semantic bridge connecting the label spaces of the two tasks. During training and inference, given the same input image, Branch~2 generates multi-level predictions based on its original LCLU taxonomy. Simultaneously, Branch~1 produces hierarchical outputs aligned with the constructed cross-domain hierarchy.

Since the two tasks share categories or exhibit semantic overlap at higher semantic levels (e.g., vegetation or cropland), we can impose a hierarchical consistency constraint on Branch~1 using the high-level predictions from Branch~2. As illustrated in Fig.~\ref{fig-overall_arch} (d), CDSA establishes inter-branch interaction through shared high-level semantics. This design not only enhances semantic alignment but also provides effective supervision from the strong LCLU model during the early training stages, leading to faster convergence and improved generalization on the target task.

Taking the transfer to a crop classification task as an example, Fig.~\ref{fig-overall_arch} (b) and (e) illustrate the complete adaptation process. We begin by establishing a cross-domain hierarchy based on prior knowledge: Vegetation $\rightarrow$ Cropland $\rightarrow$ {Rice, Maize, Soybean}. Here, ``Vegetation'' and ``Cropland'' correspond to the $L1$ and $L2$ categories in the original task, while specific crops represent the newly introduced $L3$ categories.

During the forward pass, TransLU first obtains the multi-granularity prediction score maps from the frozen Branch~2, denoted as $F^{L1}_\text{out}$, $F^{L2}_\text{out}$, and $F^{L3}_\text{out}$.
According to the cross-domain tree-structured hierarchy, we select the channels corresponding to the shared high-level concepts, i.e., vegetation at level $L1$ and cropland at level $L2$, yielding $F^{L1:\text{veg}}_\text{out}$ and $F^{L2:\text{crop}}_\text{out}$.
We then apply a softmax operation along the channel dimension to obtain soft region of interest (ROI) masks $F^{L1:\text{veg}}_\text{soft}$ and $F^{L2:\text{crop}}_\text{soft}$, which serve as spatial semantic priors distilled from the large-scale LCLU pretraining.

Meanwhile, the multi-scale features from the Branch~1 encoder are decoded to produce hierarchical feature inputs $Z^{L1}_\text{in}$, $Z^{L2}_\text{in}$, and $Z^{L3}_\text{in}$. To enforce cross-domain semantic alignment, we use the soft ROI masks to gate the corresponding category-specific features in Branch~1: $F^{L1:\text{veg}}_\text{soft}$ modulates $Z^{L1:\text{veg}}_\text{in}$ and $F^{L2:\text{crop}}_\text{soft}$ modulates $Z^{L2:\text{crop}}_\text{in}$ via element-wise multiplication, thereby emphasizing spatial regions with high confidence for vegetation and cropland according to the frozen LCLU branch.

Conversely, the remaining channels at each level ``other'' are modulated by the complementary soft ROI masks to suppress task-irrelevant signals in the target regions. The resulting gated features $\widetilde{Z}^{L1}_\text{in}$ and $\widetilde{Z}^{L2}_\text{in}$, together with $Z^{L3}_\text{in}$, are fed into the BHCCM of Branch~1 for final hierarchical prediction. For Branch~1, the inputs to BHCCM are defined as:
\begin{align}
    \widetilde{Z}_{\text{in}}^{L1:\text{veg}} &= Z_{\text{in}}^{L1:\text{veg}} \cdot F_{\text{soft}}^{L1:\text{veg}} \\
    \widetilde{Z}_{\text{in}}^{L1:\text{other}} &= Z_{\text{in}}^{L1:\text{other}} \cdot (1 - F_{\text{soft}}^{L1:\text{veg}}) \\
    \widetilde{Z}_{\text{in}}^{L2:\text{crop}} &= Z_{\text{in}}^{L2:\text{crop}} \cdot F_{\text{soft}}^{L2:\text{crop}}  \\
    \widetilde{Z}_{\text{in}}^{L2:\text{other}} &= Z_{\text{in}}^{L2:\text{other}} \cdot (1 -F_{\text{soft}}^{L2:\text{crop}})  \\
    \widetilde{Z}_{\text{in}}^{L1} &= \left[ \widetilde{Z}_{\text{in}}^{L1:\text{other}},\ \widetilde{Z}_{\text{in}}^{L1:\text{veg}} \right] \\
    \widetilde{Z}_{\text{in}}^{L2} &= \left[ \widetilde{Z}_{\text{in}}^{L2:\text{other}},\ \widetilde{Z}_{\text{in}}^{L2:\text{crop}} \right] \\
    \widetilde{Z}_{\text{in}}^{L3} &= Z_{\text{in}}^{L3}
\end{align}

To train the BHCCM in Branch~1, we employ a hybrid supervision strategy that combines ground truth with cross-branch pseudo-labels. Specifically, the fine-grained level $L3$ is supervised directly by the pixel-wise annotations from the target dataset. In contrast, addressing the lack of explicit hierarchical annotations in the target dataset, the supervision for coarse levels ($L1$ and $L2$) leverages domain knowledge transferred from Branch~2; here, the binary segmentation masks for ``Vegetation'' and ``Cropland'' predicted by Branch~2 serve as the pseudo-ground truth targets for $L1$ and $L2$, respectively. Consequently, Branch~1 is optimized to align its coarse-level predictions with the domain-generic knowledge from Branch~2, while refining its fine-grained predictions to fit the specific categories of the target task.

Overall, CDSA transfers robust high-level semantic knowledge learned by the hierarchical LCLU model to the target task by providing spatially localized semantic guidance for Branch~1, which stabilizes optimization and improves cross-domain generalization. We will validate the effectiveness of TransLU in Section~\ref{section:Experiments and Analysis} by transferring the hierarchical LCLU classification model trained on the MM-5B dataset to crop-type classification and an LCLU task with a different label taxonomy.

\subsection{Hierarchical Semantic Consistency Loss Function}
While BHCCM facilitates feature interaction across granularities, applying independent supervision at each level fails to guarantee pixel-wise semantic consistency explicitly. To address this, we propose the hierarchical semantic consistency loss ($\mathcal{L}_{\text{HSC}}$), which unifies a level-wise hierarchical cross-entropy loss ($\mathcal{L}_{\text{HCE}}$) and a structural hierarchical path consistency constraint loss ($\mathcal{L}_{\text{HPC}}$).

We first supervise classification accuracy at each hierarchical level. For the three output predictions $F_{\text{out}}^{(l)}$ ($l=1,2,3$) produced by BHCCM, we apply separate pixel-wise cross-entropy losses, aggregated to form $\mathcal{L}_{\text{HCE}}$. The loss is formulated as:
\begin{align}
    \mathcal{L}_{\text{CE}}^{(l)} &= -\frac{1}{N} \sum_{n=1}^{N} \sum_{c=1}^{C^{(l)}} \hat{y}_{n,l}^{(c)} \log p_{n,l}^{(c)} \\
    \mathcal{L}_{\text{HCE}} &= \sum_{l=1}^{3} \lambda_l \mathcal{L}_{\text{CE}}^{(l)} 
\end{align}
where $N$ is the total number of pixels, $C^{(l)}$ is the number of categories at level $l$, and $\hat{y}_{n,l}^{(c)} \in \{0,1\}$ denotes the one-hot ground-truth label, $p_{n,l}^{(c)}$ is the predicted probability of pixel $n$ belonging to class $c$ at level $l$. The hyperparameter $\lambda_l$ balances the contribution of each hierarchical level ($\lambda_l = 1$ by default).

To further regularize the global hierarchical structure, we introduce a hierarchical path consistency constraint $\mathcal{L}_{\text{HPC}}$ that explicitly encourages the predicted outputs to comply with predefined tree-structured semantic paths. We define a set of valid semantic paths conforming to the hierarchical label structure. Each path is represented as
\begin{equation}
t = \big(c^{(1)}, c^{(2)}, c^{(3)}\big)
\end{equation}
where $c^{(l)}$ denotes the category at the $l$-th hierarchical level. The set of all valid semantic paths is denoted as $\mathcal{T}$.

For each pixel $n$, the model outputs a categorical probability distribution $p_{n,l}^{(c)}$ at each hierarchical level. For a valid path $t \in \mathcal{T}$, we compute a path score by aggregating the predicted probabilities in log-probability space:
\begin{equation}
s_n(t) = \sum_{l=1}^{3} \log p_{n,l}^{\big(c^{(l)}\big)}.
\label{eq:path_score}
\end{equation}

The path scores are normalized over the valid path set $\mathcal{T}$ using a softmax operation to obtain a path-level probability distribution:
\begin{equation}
P_n(t) = \frac{\exp\big(s_n(t)\big)}{\sum_{t' \in \mathcal{T}} \exp\big(s_n(t')\big)}, 
\quad t \in \mathcal{T}.
\end{equation}

For pixel $n$, the ground-truth labels uniquely correspond to a valid semantic path.
\begin{equation}
    t_n^{gt} = \big(\hat{c}^{(1)}, \hat{c}^{(2)}, \hat{c}^{(3)}\big) \in \mathcal{T}.
    \end{equation}
    The corresponding ground-truth path distribution $\hat{P}_n(t)$ is defined as a one-hot distribution over $\mathcal{T}$:
    \begin{equation}
    \hat{P}_n(t) =
    \begin{cases}
    1, & t = t_n^{gt}, \\
    0, & \text{otherwise}.
    \end{cases}
    \end{equation}

The loss $\mathcal{L}_{\text{HPC}}$ is then formulated as the Kullback--Leibler (KL) divergence between the predicted and ground-truth path distributions:
\begin{equation}
    \mathcal{L}_{\text{HPC}} =
    \frac{1}{N} \sum_{n=1}^{N}
    \mathrm{KL}\!\left(\hat{P}_n \,\|\, P_n\right). 
\end{equation}
This loss penalizes insufficient probability mass assigned to the ground-truth semantic path, thereby encouraging structurally consistent predictions. It is worth noting that since the ground-truth path distribution $\hat{P}_n$ is deterministic in MM-5B (i.e., a one-hot vector), minimizing this KL divergence is mathematically equivalent to minimizing the standard cross-entropy loss. However, we formulate it as a distribution matching problem to maintain theoretical generality. This formulation offers greater flexibility by explicitly encouraging the alignment between the predicted and target distributions, allowing for potential extensions to soft-label supervision or uncertainty-aware learning where ground-truth paths may not be unique or deterministic.

The final training objective combines the loss $\mathcal{L}_{\text{HCE}}$ and the loss $\mathcal{L}_{\text{HPC}}$:
\begin{equation}
    \mathcal{L}_{\text{HSC}} = \mathcal{L}_{\text{HCE}} + \alpha \mathcal{L}_{\text{HPC}},
\end{equation}
\noindent
where $\alpha$ controls the relative importance of the path consistency constraint. By jointly optimizing level-wise classification accuracy and global hierarchical structure, the proposed training objective effectively enhances semantic coordination across hierarchical levels and mitigates cross-level prediction conflicts.

During inference, we further adopt a joint score-based path selection (JSPS) strategy to enforce strict hierarchical consistency. Specifically, the path score $s_n(t)$ in Eq.~\eqref{eq:path_score} is reused, and the prediction space is restricted to the valid path set $\mathcal{T}$. The optimal semantic path is selected as
\begin{equation}
t_n^{*} = \arg\max_{t \in \mathcal{T}} s_n(t),
\end{equation}
and the final prediction at each hierarchical level is obtained from the selected path. In this way, JSPS complements the training-time path consistency constraint by enforcing strict hierarchical consistency during inference.

\section{Experiments}
\label{section:Experiments and Analysis}
In this section, we present comprehensive experiments on three benchmark datasets to validate the effectiveness of the proposed HieraRS.

\subsection{Datasets}
1) \textit{\textbf{MM-5B:}} We evaluate BHCCM on the MM-5B dataset for hierarchical LCLU classification. We used 107 images for training and 29 for validation. Sentinel-2 and GaoFen-2 images were cropped into non-overlapping patches of $512 \times 512$ and $640 \times 640$, respectively. To reduce computational overhead, 1-m Google Earth imagery was downsampled to 2~m and cropped to $896 \times 896$. This yields training/validation sample counts of 2,370\,/\,632 for Sentinel-2, 15,091\,/\,4,110 for GaoFen-2, and 28,679\,/\,7,635 for Google Earth. As shown in Fig.~\ref{fig-MM-5B-hiera-structure}, the annotations follow a three-level hierarchy with 4, 9, and 18 categories at $L1$, $L2$, and $L3$, respectively.

\begin{figure}[tb]
    \centering 
    \centerline{\includegraphics[width=0.80\linewidth]{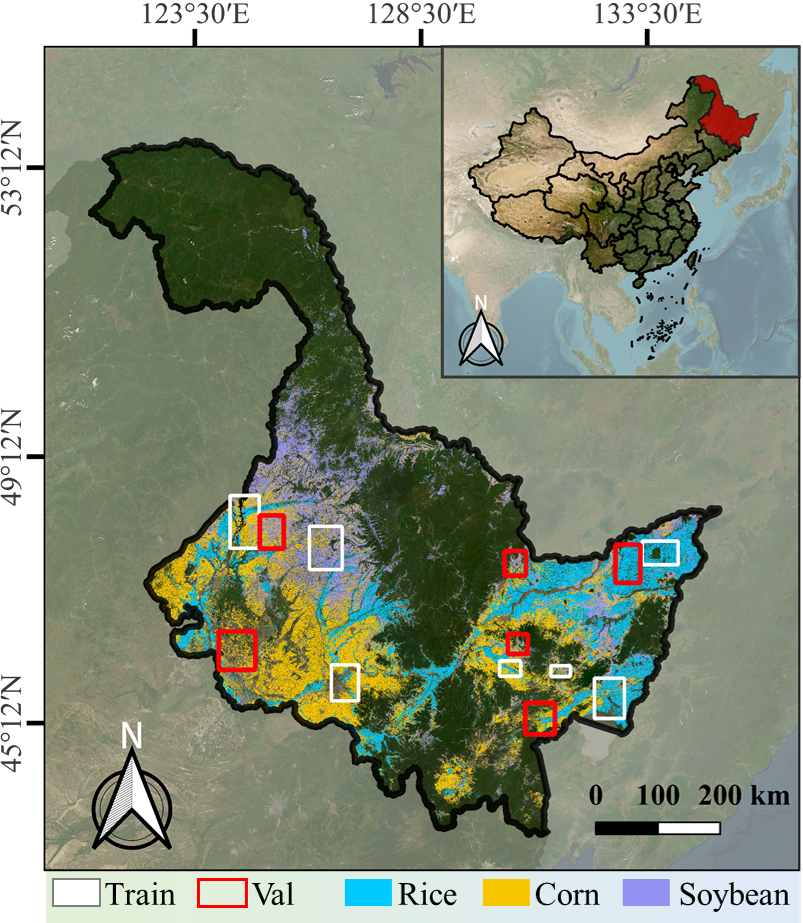}}
    \caption{Illustration of the Crop10m study area in Heilongjiang Province, Northeast China. The dataset focuses on three crop categories: rice, corn, and soybean. The training and validation regions are delineated by white and red bounding boxes, respectively.}
    \label{fig-crop10m-area}
\end{figure}

2) \textit{\textbf{Crop10m:}} To validate TransLU in a cross-domain scenario transferring from hierarchical LCLU to crop classification, we adopt the ground-truth labels of 2019 from You et al.~\cite{crop10m_dataset_2021}. Accordingly, we acquired spatially aligned Sentinel-2 imagery from the corresponding period covering the primary crop regions (rice, maize, and soybean) in Heilongjiang, China. As illustrated in Fig.~\ref{fig-crop10m-area}, the study area is spatially partitioned into independent training and validation regions. The imagery is cropped into $512 \times 512$ patches, resulting in 678 training samples and 296 validation samples. The specific cross-domain hierarchical structure is detailed in Fig.~\ref{fig-overall_arch}(e).

3) \textit{\textbf{WHDLD:}} To demonstrate the adaptation of TransLU to a distinct hierarchy, we employ the WHDLD dataset containing six annotated classes. It comprises 4,940 RGB images ($256 \times 256$), split by a 2:1 ratio into 3,294 training and 1,646 validation samples. As illustrated in Fig.~\ref{WHDLD-result}, the categories are organized into a two-level hierarchical structure.

\subsection{Experimental Setup}
To assess the versatility of BHCCM across architectures, we benchmark both CNN- and Transformer-based models on the MM-5B dataset, comparing flat classification baselines against our hierarchical approach.  Specifically, DeepLabv3+~\cite{deeplabv3plus_2018_ECCV} utilizes the SGD optimizer, whereas ConvNeXt~\cite{convnext_2022_cvpr}, SegNext~\cite{segnext_2022_NeurIPS}, DeiT3~\cite{deit3_2022_ECCV}, and Swin~\cite{swin_2021_ICCV} employ the AdamW optimizer. All models are implemented in MMSegmentation~\cite{mmseg2020} with a batch size of 8 on four NVIDIA A800 GPUs and are trained for 80k iterations for each data source. Performance is evaluated via mean Intersection over Union (mIoU). For cross-domain transfer experiments on Crop10m and WHDLD, settings remain consistent, except the training duration is reduced to 20k iterations.

\subsection{Hierarchical LCLU Classification on MM-5B}

Tables~\ref{GF2-Hiera-results-table}, \ref{S2-Hiera-results-table}, and \ref{Google-Hiera-results-table} report the hierarchical LCLU classification results on the three data sources of MM-5B. We compare models trained with and without the proposed BHCCM. For flat baselines that do not natively support multi-level predictions, $L3$ results are mapped to $L1$ and $L2$ categories following the hierarchy in Fig.~\ref{fig-MM-5B-hiera-structure} to derive the corresponding metrics.

To further benchmark the competitiveness of our approach, we compare the proposed BHCCM against two representative state-of-the-art hierarchical models. Specifically, we selected HierU-Net~\cite{hiera-unet_2024_TGRS} as it represents an architecture-driven approach, employing a specialized dual-network design tailored for hierarchical LCLU tasks. Conversely, HSSN~\cite{hieraseg_2024_TPAMI} was chosen because it aligns with the conceptual philosophy of our BHCCM, aiming to transform existing flat models into hierarchical segmenters with minimal structural modifications. Including these diverse baselines allows for a comprehensive evaluation against both specialized architectures and general-purpose hierarchical adaptation frameworks.

\subsubsection{Comparison with Flat Baselines on GaoFen-2}
As presented in Table~\ref{GF2-Hiera-results-table}, integrating BHCCM consistently improves classification accuracy over baselines across diverse backbones and hierarchical levels on the GaoFen-2 data source of the MM-5B dataset. Specifically, representative CNN-based models exhibit robust gains: DeepLabv3+ with ResNet-101 enhanced by BHCCM achieves a 1.80\% mIoU increase at the $L3$ level, while ConvNeXt-B and ConvNeXt-L show improvements of 0.77\% and 0.86\%, respectively. Most notably, BHCCM significantly empowers lightweight models. SegNeXt-B improves by 0.98\% at $L3$, and impressively, the smaller SegNeXt-S with BHCCM (74.77\%) surpasses even the larger SegNeXt-B baseline (74.74\%), highlighting exceptional efficiency.

Transformer-based architectures demonstrate similarly substantial gains. DeiT3-S achieves a remarkable 2.01\% improvement at L3, outperforming the larger DeiT3-B baseline. Meanwhile, DeiT3-B and DeiT3-L improve by 1.15\% and 0.63\%, respectively, with DeiT3-S also showing consistent gains at $L2$ and $L1$. Overall, the proposed method yields robust enhancements across the hierarchy. While Swin-L with BHCCM secures optimal results at coarser levels, SegNeXt-L combined with BHCCM achieves the highest accuracy of 76.59\% at the most challenging $L3$ level.

% MM-5B GaoFen-2
\begin{table}[t]
    \centering
    \caption{mIoU Performance Comparison on the GaoFen-2 Source of MM-5B. \textbf{\textcolor{red}{Red}}/\textbf{\textcolor{blue}{Blue}} Highlights Optimal/Sub-Optimal Values.}
    \label{GF2-Hiera-results-table}
    \renewcommand\arraystretch{1.1} 
    \resizebox{1.0\linewidth}{!}{ 
    \begin{tabular}{c|l|l|c c c}
        \toprule
        \textbf{Method} & \textbf{Backbone} & \textbf{Decoder} & $\bm{L3}$ & $\bm{L2}$ & $\bm{L1}$   \\   
        \midrule  
        % ================= BASELINE SECTION =================
        \multirow{12}{*}{{\shortstack{Flat\\Baseline}}}
        % ResNet
         & \multirow{2}{*}{ResNet-101\cite{resnet_2016_cvpr}}      & DeepLabv3+\cite{deeplabv3plus_2018_ECCV}     & 70.59   & 81.22  & 93.16  \\
         &                                                         & HierU-Net (single)\cite{hiera-unet_2024_TGRS}  & 68.74   & 79.65  & 92.10  \\
         \cline{2-6} 
        % MSCA 
         & MSCA-S\cite{segnext_2022_NeurIPS}      & \multirow{3}{*}{SegNeXt\cite{segnext_2022_NeurIPS}}  & 74.29   & 82.32  & 94.39 \\
         & MSCA-B                                 &                                                      & 74.74   & 83.37  & 94.59 \\
         & MSCA-L                                 &                                                      & 76.23   & 83.82  & 94.85 \\
         \cline{2-6}
        % ConvNeXt 
         & ConvNeXt-B\cite{convnext_2022_cvpr}    & \multirow{2}{*}{UperNet\cite{upernet_2018_ECCV}}  & 73.39   & 82.38 & 93.73  \\
         & ConvNeXt-L                             &                                                   & 75.32   & 83.78 & 94.47  \\
         \cline{2-6}
        % Swin 
         & Swin-B\cite{swin_2021_ICCV}            & \multirow{2}{*}{UperNet}                          & 75.10   & 84.08 & 94.73  \\
         & Swin-L                                 &                                                   & 75.67   & 84.11 & 94.85  \\
         \cline{2-6}
        % DeiT3 
         & DeiT3-S\cite{deit3_2022_ECCV}          & \multirow{3}{*}{UperNet}                          & 71.46   & 81.46 & 93.76  \\
         & DeiT3-B                                &                                                   & 72.72   & 82.17 & 94.00  \\
         & DeiT3-L                                &                                                   & 75.15   & 83.63 & 94.35  \\       
        % ================= HierU-Net SECTION =================
        \midrule
        HierU-Net\cite{hiera-unet_2024_TGRS}     
        & ResNet-101                             & HierU-Net (dual U-Net)                             & 69.25   & 79.86  & ---    \\ 
        \midrule
        % ================= HSSN SECTION =================
        \multirow{3}{*}{HSSN\cite{hieraseg_2024_TPAMI}}
        & ResNet-101                             & DeepLabv3+                                         & 71.35   & 81.57  & 93.26  \\
        & ConvNeXt-B                             & UperNet                                            & 73.86   & 82.73  & 94.01  \\
        & Swin-B                                 & UperNet                                            & 75.34   & 84.20  & 94.87  \\
        \midrule
        % ================= BHCCM (OURS) SECTION =================
        \multirow{11}{*}{\textbf{\shortstack{BHCCM\\(Ours)}}}
        % ResNet
        & \multirow{2}{*}{ResNet-101}           & DeepLabv3p\,+\,BHCCM                                & 72.39   & 82.33  & 93.67  \\
        &                                       & HierU-Net (single)\,+\,BHCCM                                 & 69.78   & 80.13  & 92.36  \\
        \cline{2-6}
        % MSCA 
        & MSCA-S                                & \multirow{3}{*}{SegNeXt\,+\,BHCCM}                  & 74.77   & 83.57 & 94.53 \\
        & MSCA-B                                &                                                     & 75.72   & 83.63 & 94.62 \\
        & MSCA-L                                &                                                     & \textbf{\textcolor{red}{76.59}}   & 84.32 & 94.86 \\
        \cline{2-6}
        % ConvNeXt 
        & ConvNeXt-B                            & \multirow{2}{*}{UperNet\,+\,BHCCM}                  & 74.16   & 82.84 & 94.22 \\
        & ConvNeXt-L                            &                                                     & \textbf{\textcolor{blue}{76.18}}   & 83.95 & 94.81 \\  
        \cline{2-6}
        % Swin 
        & Swin-B                                & \multirow{2}{*}{UperNet\,+\,BHCCM}                  & 75.62   & \textbf{\textcolor{blue}{84.36}} & \textbf{\textcolor{blue}{95.08}} \\
        & Swin-L                                &                                                     & 76.14   & \textbf{\textcolor{red}{84.52}} & \textbf{\textcolor{red}{95.16}} \\
        \cline{2-6}
        % DeiT3 
        & DeiT3-S                               & \multirow{3}{*}{UperNet\,+\,BHCCM}                  & 73.47   & 82.55 & 94.39 \\
        & DeiT3-B                               &                                                     & 73.87   & 82.49 & 94.41 \\
        & DeiT3-L                               &                                                     & 75.78   & 84.11 & 94.71 \\
    \bottomrule
    \end{tabular}
    }
\end{table}

\subsubsection{Comparison with Flat Baselines on Sentinel-2}
Table~\ref{S2-Hiera-results-table} presents the results on the Sentinel-2 data source. Due to the lower spatial resolution compared to Gaofen-2, fine-grained classification becomes significantly more challenging, resulting in lower baselines. However, BHCCM demonstrates exceptional robustness in this difficult scenario, with gains being particularly pronounced at the most complex $L3$ level. For instance, DeepLabv3+ sees a substantial 1.72\% surge at $L3$, while also maintaining significant improvements at the $L2$ and $L1$ levels. Performance across modern backbones further validates this trend. The CNN-based ConvNeXt-L with BHCCM exhibits strong competitiveness, achieving 72.65\% at $L2$ and 84.22\% at $L1$, marking the best and second-best results among all comparisons. Similarly, the lightweight SegNeXt-S with BHCCM (62.27\%) efficiently outperforms the heavier SegNeXt-B baseline. Ultimately, the Transformer-based Swin-L integrated with BHCCM attains the highest accuracy of 64.64\% at $L3$. These results underscore the versatility of BHCCM across diverse architectures and its efficacy on multi-spectral imagery with moderate spatial resolution.

% MM-5B Sentinel-2
\begin{table}[t]
    \centering
    \caption{mIoU Performance Comparison on the Sentinel-2 Source of MM-5B. \textbf{\textcolor{red}{Red}}/\textbf{\textcolor{blue}{Blue}} Highlights Optimal/Sub-Optimal Values.}
    \label{S2-Hiera-results-table}
    \renewcommand\arraystretch{1.1} 
    \resizebox{1.0\linewidth}{!}{ 
        \begin{tabular}{c|l|l|c c c}
        \toprule
        \textbf{Method} & \textbf{Backbone} & \textbf{Decoder} & $\bm{L3}$ & $\bm{L2}$ & $\bm{L1}$   \\   
        \midrule
        % ================= BASELINE SECTION =================
        \multirow{12}{*}{{\shortstack{Flat\\Baseline}}}
        % ResNet 
         & \multirow{2}{*}{ResNet-101\cite{resnet_2016_cvpr}}      & DeepLabv3+\cite{deeplabv3plus_2018_ECCV}    & 59.02   & 69.06  & 82.06  \\
         &                                                         & HierU-Net (single)\cite{hiera-unet_2024_TGRS} & 58.73   & 68.74  & 81.79  \\
         \cline{2-6}
        % MSCA 
         & MSCA-S\cite{segnext_2022_NeurIPS}      & \multirow{3}{*}{SegNeXt\cite{segnext_2022_NeurIPS}}  & 61.54   & 70.66  & 82.77  \\
         & MSCA-B                                 &                                                      & 62.20   & 71.25  & 83.04  \\
         & MSCA-L                                 &                                                      & 63.15   & 71.93  & 83.59  \\
         \cline{2-6}
        % ConvNeXt 
         & ConvNeXt-B\cite{convnext_2022_cvpr}    & \multirow{2}{*}{UperNet\cite{upernet_2018_ECCV}}    & 61.57    & 70.90 & 82.50  \\
         & ConvNeXt-L                             &                                                     & 63.16    & 72.27 & 83.87  \\
         \cline{2-6}
        % Swin 
         & Swin-B\cite{swin_2021_ICCV}            & \multirow{2}{*}{UperNet}                            & 63.20    & 72.12 & 84.05  \\
         & Swin-L                                 &                                                     & 64.16    & 72.30 & 84.13  \\
         \cline{2-6}
        % DeiT3 
         & DeiT3-S\cite{deit3_2022_ECCV}          & \multirow{3}{*}{UperNet}                            & 58.09    & 67.35 & 80.28  \\
         & DeiT3-B                                &                                                     & 60.50    & 69.44 & 81.02  \\
         & DeiT3-L                                &                                                     & 61.47    & 71.12 & 82.29  \\           
        % ================= HierU-Net SECTION =================
        \midrule
        HierU-Net\cite{hiera-unet_2024_TGRS}     
        & ResNet-101                             & HierU-Net (dual U-Net)                              & 59.77   & 68.95  & ---    \\ 
        \midrule
        % ================= HSSN SECTION =================
        \multirow{3}{*}{HSSN\cite{hieraseg_2024_TPAMI}}
        & ResNet-101                             & DeepLabv3+                                           & 60.13   & 69.60  & 82.54   \\
        & ConvNeXt-B                             & UperNet                                              & 61.93   & 71.15  & 82.67   \\
        & Swin-B                                 & UperNet                                              & 63.49   & 72.23  & 84.17   \\
        \midrule
        % ================= BHCCM (OURS) SECTION =================
        \multirow{11}{*}{\textbf{\shortstack{BHCCM\\(Ours)}}}
        % ResNet 
        & \multirow{2}{*}{ResNet-101}           & DeepLabv3p\,+\,BHCCM                                 & 60.74   & 70.12  & 82.97  \\
        &                                       & HierU-Net (single)\,+\,BHCCM                           & 60.11   & 69.58  & 82.34  \\
        \cline{2-6}
        % MSCA 
        & MSCA-S                                & \multirow{3}{*}{SegNeXt + BHCCM}                     & 62.27   & 71.58 & 82.95   \\
        & MSCA-B                                &                                                      & 63.14   & 71.90 & 83.30   \\
        & MSCA-L                                &                                                      & 63.47   & 72.44 & 83.74   \\
        \cline{2-6}
        % ConvNeXt 
        & ConvNeXt-B                            & \multirow{2}{*}{UperNet + BHCCM}                     & 62.16    & 71.38 & 82.81 \\
        & ConvNeXt-L                            &                                                      & 63.62    & \textbf{\textcolor{red}{72.65}} & \textbf{\textcolor{blue}{84.22}} \\  
        \cline{2-6}
        % Swin  
        & Swin-B                                & \multirow{2}{*}{UperNet + BHCCM}                     & \textbf{\textcolor{blue}{63.87}}   & 72.36 & 84.20 \\
        & Swin-L                                &                                                      & \textbf{\textcolor{red}{64.64}}    & \textbf{\textcolor{blue}{72.61}} & \textbf{\textcolor{red}{84.39}} \\
        \cline{2-6}
        % DeiT3 
        & DeiT3-S                               & \multirow{3}{*}{UperNet + BHCCM}                     & 58.73    & 67.73 & 80.60 \\
        & DeiT3-B                               &                                                      & 61.08    & 69.85 & 81.31 \\
        & DeiT3-L                               &                                                      & 61.89    & 71.48 & 82.54 \\
        \bottomrule
        \end{tabular}
    }
\end{table}

\subsubsection{Comparison with Flat Baselines on Google Earth}
Table~\ref{Google-Hiera-results-table} details the quantitative results on the Google Earth source. This dataset presents distinct challenges due to the high intra-class variability arising from high spatial resolution and the insufficient spectral information of RGB-only spectra. Despite these constraints, integrating BHCCM yields substantial improvements across various backbones. Notably, the classic DeepLabv3+ achieves remarkable mIoU gains of 4.23\%, 3.86\%, and 3.21\% across the $L3$, $L2$, and $L1$ levels, respectively. Furthermore, advanced architectures demonstrate strong adaptability; for instance, SegNeXt-L combined with BHCCM attains the highest accuracy at the fine-grained levels (61.05\% at $L3$ and 66.04\% at $L2$), while Swin-L secures the optimal performance at the coarse $L1$ level (74.21\%). Consistently, all other models integrated with BHCCM significantly outperform their respective baselines. These results validate the adaptability of the proposed BHCCM to high-resolution imagery with limited spectral information.

% MM-5B Google Earth
\begin{table}[tb]
    \centering
    \caption{mIoU Performance Comparison on the Google Earth Source. \textbf{\textcolor{red}{Red}}/\textbf{\textcolor{blue}{Blue}} Highlights Optimal/Sub-Optimal Values.}
    \label{Google-Hiera-results-table}
    \renewcommand\arraystretch{1.1}
    \resizebox{1.0\linewidth}{!}{ 
        \begin{tabular}{c|l|l|c c c}
        \toprule
        \textbf{Method} & \textbf{Backbone} & \textbf{Decoder} & $\bm{L3}$ & $\bm{L2}$ & $\bm{L1}$   \\   
        \midrule
        % ================= BASELINE SECTION =================
        \multirow{12}{*}{{\shortstack{Flat\\Baseline}}}
        % ResNet 
         & \multirow{2}{*}{ResNet-101\cite{resnet_2016_cvpr}}      & DeepLabv3+\cite{deeplabv3plus_2018_ECCV}        & 55.27   & 59.26  & 68.60  \\
         &                                                         & HierU-Net (single)\cite{hiera-unet_2024_TGRS}     & 54.15   & 58.94  & 68.28  \\
         \cline{2-6}
        % MSCA 
         & MSCA-S\cite{segnext_2022_NeurIPS}      & \multirow{3}{*}{SegNeXt\cite{segnext_2022_NeurIPS}} & 57.72    & 62.82  & 70.87 \\
         & MSCA-B                                 &                                                     & 57.81    & 63.65  & 71.53 \\
         & MSCA-L                                 &                                                     & 58.87    & 64.18  & 71.96 \\
         \cline{2-6}
        % ConvNeXt 
         & ConvNeXt-B\cite{convnext_2022_cvpr}    & \multirow{2}{*}{UperNet\cite{upernet_2018_ECCV}}    & 57.37    & 62.53 & 71.27 \\
         & ConvNeXt-L                             &                                                     & 58.54    & 63.42 & 71.50 \\
         \cline{2-6}
        % Swin 
         & Swin-B\cite{swin_2021_ICCV}            & \multirow{2}{*}{UperNet}                            & 58.72    & 63.57 & 71.44 \\
         & Swin-L                                 &                                                     & 59.93    & 64.15 & 72.63 \\
         \cline{2-6}
        % DeiT3 
         & DeiT3-S\cite{deit3_2022_ECCV}          & \multirow{3}{*}{UperNet}                            & 55.68    & 62.01 & 70.81 \\
         & DeiT3-B                                &                                                     & 57.34    & 63.00 & 71.17 \\
         & DeiT3-L                                &                                                     & 58.23    & 64.11 & 71.77 \\           
        % ================= HierU-Net SECTION =================
        \midrule
        HierU-Net\cite{hiera-unet_2024_TGRS}     
        & ResNet-101                             & HierU-Net (dual U-Net)                              & 55.34    & 59.37  & ---   \\ 
        \midrule
        % ================= HSSN SECTION =================
        \multirow{3}{*}{HSSN\cite{hieraseg_2024_TPAMI}}
        & ResNet-101                             & DeepLabv3+                                           & 57.42   & 61.33 & 70.06  \\
        & ConvNeXt-B                             & UperNet                                              & 58.41   & 63.34 & 71.90  \\
        & Swin-B                                 & UperNet                                              & 59.32   & 63.94 & 71.98  \\
        \midrule
        % ================= BHCCM (OURS) SECTION =================
        \multirow{11}{*}{\textbf{\shortstack{BHCCM\\(Ours)}}}
        % ResNet 
        & \multirow{2}{*}{ResNet-101}           & DeepLabv3p\,+\,BHCCM                                 & 59.50   & 63.12 & 71.81  \\
        &                                       & HierU-Net (single)\,+\,BHCCM                         
        
        & 57.61   & 61.24 & 69.67  \\
        \cline{2-6}
        % MSCA 
        & MSCA-S                               & \multirow{3}{*}{SegNeXt + BHCCM}                      & 60.07   & 64.90 & 73.13  \\
        & MSCA-B                               &                                                       & 60.95   & 65.79 & 73.86  \\
        & MSCA-L                               &                                                       & \textbf{\textcolor{red}{61.05}}    & \textbf{\textcolor{red}{66.04}}  & \textbf{\textcolor{blue}{74.11}} \\
        \cline{2-6}
        % ConvNeXt 
        & ConvNeXt-B                           & \multirow{2}{*}{UperNet + BHCCM}                      & 59.98   & 64.57 & 73.15  \\
        & ConvNeXt-L                           &                                                       & 60.68   & 65.05 & 73.51  \\  
        \cline{2-6}
        % Swin 
        & Swin-B                               & \multirow{2}{*}{UperNet + BHCCM}                      & 60.24   & 64.83 & 73.05  \\
        & Swin-L                               &                                                       & \textbf{\textcolor{blue}{61.03}}    &  \textbf{\textcolor{blue}{65.13}} & \textbf{\textcolor{red}{74.21}} \\
        \cline{2-6}
        % DeiT3 
        & DeiT3-S                              & \multirow{3}{*}{UperNet + BHCCM}                      & 57.83   & 63.78 & 72.33 \\
        & DeiT3-B                              &                                                       & 59.30   & 64.57 & 72.28 \\
        & DeiT3-L                              &                                                       & 59.71   & 65.00 & 72.75 \\
        \bottomrule
        \end{tabular}
    }
\end{table}

\subsubsection{Comparison with SOTA Hierarchical Methods}
To further validate the superiority of our approach, beyond the comparison with flat baselines, we benchmark BHCCM against two representative hierarchical segmentation methods reported in the middle sections of Tables~\ref{GF2-Hiera-results-table}--\ref{Google-Hiera-results-table}: HierU-Net~\cite{hiera-unet_2024_TGRS}, an architecture-driven dual-network approach, and HSSN~\cite{hieraseg_2024_TPAMI}, a multi-label classification framework.

We first benchmark our approach against HierU-Net, which employs a specific dual-network architecture to perform coarse-to-fine segmentation, where the coarse-level output constrains the fine-level prediction. To ensure a fair comparison, we established a controlled experimental setup. Specifically, we implemented a flat baseline by equipping a ResNet-101 backbone with a single-stream HierU-Net decoder. This stands in contrast to the original HierU-Net, which relies on a dual-network cascaded structure to enforce coarse-to-fine constraints. Our proposed model is then constructed by integrating the BHCCM module into the flat baseline. It is worth noting that the original HierU-Net is rigidly designed to predict only two hierarchical levels ($L2$ and $L3$), and thus cannot directly produce $L1$-level metrics. On GaoFen-2 (Table~\ref{GF2-Hiera-results-table}), our method not only outperforms the original cascaded HierU-Net with a 0.53\% increase at the fine-grained $L3$ level, but also secures a gain of 0.27\% at the intermediate $L2$ level. Similarly, on Sentinel-2 (Table~\ref{S2-Hiera-results-table}), our approach maintains its advantage, improving the mIoU by 0.34\% at $L3$ and 0.63\% at $L2$ compared to the original model. The improvements are most pronounced on Google Earth (Table~\ref{Google-Hiera-results-table}). Our method significantly surpasses the HierU-Net by notable margins of 2.27\% at $L3$ and 1.87\% at $L2$.

This demonstrates that our single-stream framework equipped with BHCCM is more effective across all hierarchical levels than the cascaded dual-network design, which relies solely on unidirectional coarse-to-fine feature transfer. Furthermore, such cascaded architectures suffer from inherent scalability issues; as the number of hierarchical levels in a dataset increases, the model complexity grows multiplicatively due to the need for stacking additional networks. In contrast, our BHCCM facilitates comprehensive multi-level feature interaction with minimal structural modifications, simultaneously enhancing prediction performance and consistency at every hierarchical level.

Subsequently, we compare BHCCM with HSSN, which formulates the hierarchical task as a pixel-wise multi-label classification problem by modifying the segmentation head to output predictions for all three levels simultaneously. Similar to our JSPS strategy, HSSN also employs a path search mechanism ensuring output consistency. However, despite this similarity, BHCCM consistently outperforms HSSN across all datasets and hierarchical levels. 

Specifically, on GaoFen-2(Table~\ref{GF2-Hiera-results-table}), comparisons using the ResNet-101 backbone show that BHCCM leads HSSN by 1.04\% at $L3$, while simultaneously achieving consistent gains of 0.76\% at $L2$ and 0.41\% at $L1$. This superiority extends to modern architectures; both ConvNeXt-B and Swin-B backbones equipped with BHCCM consistently outperform HSSN counterparts across all three levels. On Sentinel-2 (Table~\ref{S2-Hiera-results-table}), BHCCM with ResNet-101 demonstrates a clear advantage, improving the $L3$ accuracy by 0.61\% over HSSN, alongside gains of 0.52\% at $L2$ and 0.43\% at $L1$. Similarly, the ConvNeXt-B model achieves comprehensive improvements across the hierarchy. The performance gap is most pronounced on Google Earth (Table~\ref{Google-Hiera-results-table}). BHCCM exhibits comprehensive superiority across all backbones. With ResNet-101, our method surpasses HSSN by over 2\% at $L3$, and maintains a substantial lead of 1.79\% at $L2$ and 1.75\% at $L1$. Notably, the ConvNeXt-B backbone also exhibits substantial improvements, surpassing HSSN by 1.57\% at $L3$ and 1.23\% at $L2$. These comparisons validate that by explicitly enhancing information interaction across granularities via the BHCCM module, our method effectively surpasses multi-label classification paradigms that lack explicit cross-level feature interaction and rely solely on loss functions to enforce hierarchical consistency, thereby achieving consistent performance gains across different hierarchical levels.

Overall, the comparative results demonstrate that BHCCM validates its superiority over representative state-of-the-art hierarchical methods. By explicitly enhancing information interaction across different granularities via the BHCCM structure and enforcing path-level regularization through the $\mathcal{L}_{\text{HSC}}$ loss, our method effectively surpasses existing hierarchical approaches while exhibiting strong adaptability to diverse model architectures and hierarchical depths.

\begin{figure*}[tb]
    \centering 
    \centerline{\includegraphics[width=0.90\textwidth]{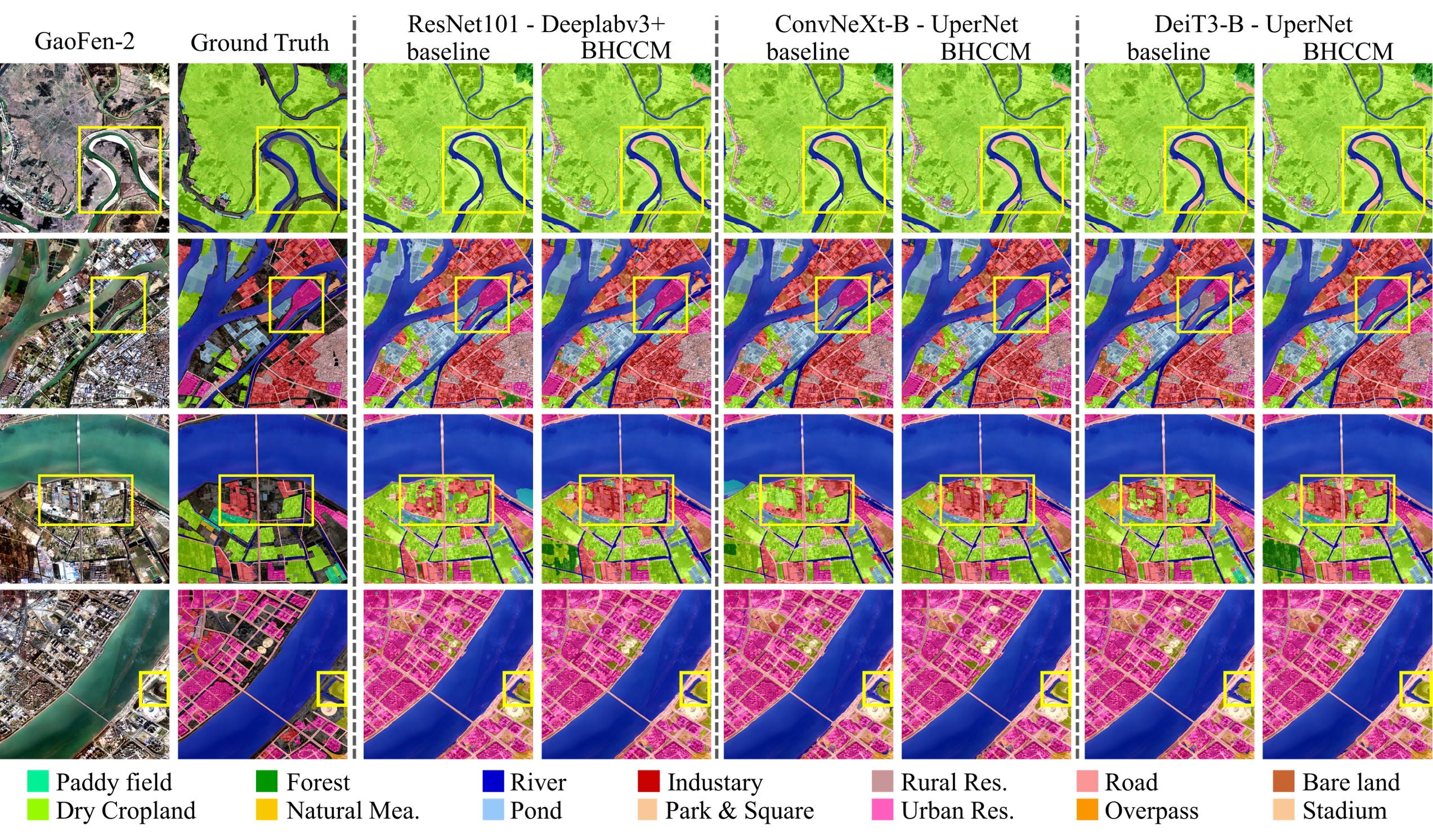}}
    \caption{Visualization results on the GaoFen-2 source of the MM-5B dataset. The figure demonstrates the performance improvements at the finest level ($L3$) achieved by the proposed BHCCM.}
    \label{fig-gf2_result}
\end{figure*}

\subsubsection{Qualitative Results}

Qualitatively, Fig.~\ref{fig-gf2_result} visualizes the L3 performance gains on GaoFen-2 data. Across representative models (DeepLabv3+, ConvNeXt, and DeiT3), integrating BHCCM yields sharper boundary delineation, particularly between urban and rural residential areas as shown in the first row. This demonstrates that BHCCM effectively enhances the discriminability of subclasses sharing the same $L2$ parent category. Moreover, gross identification errors are significantly reduced; for instance, as shown in the third row, the industrial area is rarely misclassified as dry cropland hierarchical paths at $L1$ and $L2$ are distinct. This result highlights the improvement in fine-grained recognition facilitated by the structural constraints from distinct coarse-level features. This robustness stems from the multi-level consistency constraint, which effectively suppresses cross-level semantic contradictions. Consequently, as shown in the fourth row, fine-grained details---such as the boundaries between narrow water bodies and surrounding parks---are identified more precisely due to the clear separation of inconsistent parent classes.

\begin{figure*}[tb]
    \centering 
    \centerline{\includegraphics[width=0.90\textwidth]{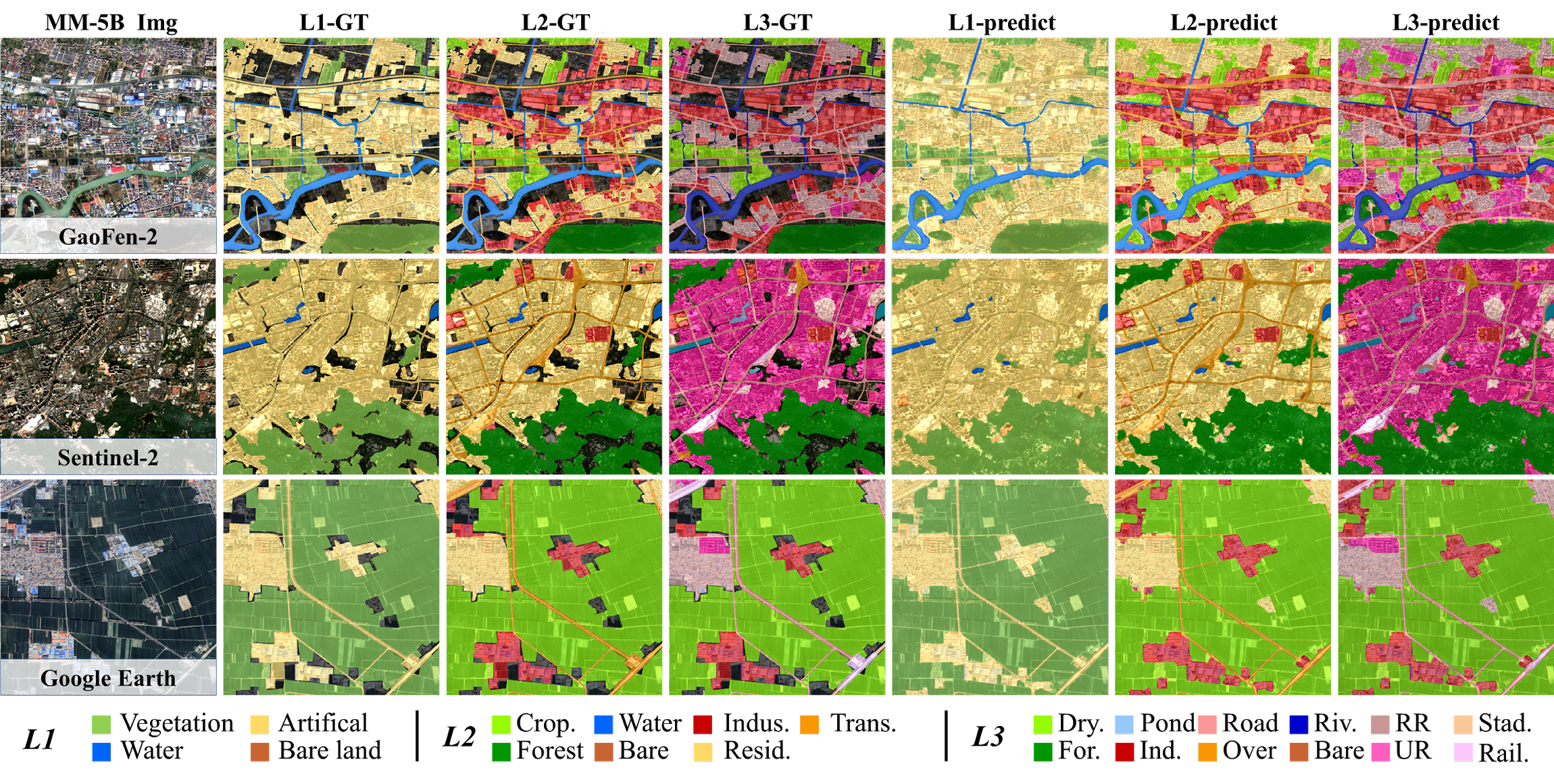}}
    \caption{Hierarchical LCLU classification results of ConvNeXt-B equipped with the proposed BHCCM across the three data sources in the MM-5B dataset. The figure displays the interpretation results for different sources, highlighting the strict consistency of hierarchical predictions across levels $L1$, $L2$, and $L3$ within each individual source.}
        \label{fig-MM-5B-Hiera}
\end{figure*}

Furthermore, Fig.~\ref{fig-MM-5B-Hiera} demonstrates the interpretation results across varying spatial resolutions. It highlights that, across all three data sources, the BHCCM-integrated model generates predictions that strictly adhere to the pre-defined tree-structured hierarchy, ensuring logical consistency from coarse to fine levels.

\subsection{Transfer Hierarchical LCLU Model to Crop Classification}
To evaluate the effectiveness of the proposed TransLU framework, we assess its transferability in crop classification scenarios using the Crop10m dataset. This experiment is designed to examine the generalization and scalability of TransLU in real-world applications. Specifically, a hierarchical LCLU classification model trained on Sentinel-2 source of the MM-5B dataset is transferred to the Crop10m dataset, enabling the expansion of crop categories while preserving the model's capacity for hierarchical LCLU classification.

\subsubsection{Quantitative Results}

Table~\ref{crop-results-table} presents the transfer learning results using the proposed TransLU framework across three representative architectures: ConvNeXt (CNN), DeiT3 (Transformer), and SegNeXt (Lightweight CNN). The hierarchical LCLU models utilized for transfer are initialized with the weights of ConvNeXt-B, DeiT3-B, and SegNeXt-B reported in Table~\ref{S2-Hiera-results-table}. By comparing the single-branch baselines against our dual-branch TransLU approach, we observe consistent and substantial performance gains across all models.

Specifically, TransLU boosts the mIoU of ConvNeXt-B by 2.69\%, achieving 80.58\%. Notable class-wise improvements are observed: 2.13\% for rice, 2.64\% for corn, and 2.33\% for soybean. Most significantly, the Others class sees a remarkable gain of 3.64\% (from 78.26\% to 81.90\%). This indicates that the LCLU prior knowledge from Branch~2 effectively assists Branch~1 in identifying non-crop background elements (e.g., water, artificial surfaces, and forest), thereby reducing false positives in crop classification. Similar trends are evident in other architectures: the Transformer-based DeiT3-B achieves a 2.66\% mIoU increase, with substantial gains of 2.21\%, 2.40\%, and 1.93\% for rice, corn, and soybean, respectively. Notably, the Others class again exhibits the most prominent improvement of 4.11\%, reinforcing the effective separation of background elements. Finally, the lightweight SegNeXt-B also benefits from TransLU, improving by 2.64\%. 

Collectively, these results demonstrate that TransLU facilitates the efficient transfer of general LCLU models to downstream tasks characterized by distinct semantic categories. Furthermore, the consistent improvements across diverse backbones validate that the CDKS and CDSA designs within TransLU are adaptable to various model architectures, effectively enhancing discriminability for novel categories.

\begin{table*}[t]
    \centering
    \caption{Performance Comparison of the Proposed TransLU Framework on the Crop10m Dataset.}
    \label{crop-results-table}
    \renewcommand\arraystretch{1.1} 
	\fontsize{7}{8}\selectfont
    \resizebox{0.95\textwidth}{!}{ 
        \begin{tabular}{l|l|l|l|l|c|c c c c}
        \toprule
        \multirow{2}{*}{\textbf{Method}} & \multicolumn{2}{c|}{\textbf{Branch1}}   & \multicolumn{2}{c|}{\textbf{Branch2}} & \multicolumn{5}{c}{\textbf{Classification Performance (IoU)}} \\  
        \cline{2-10}  
                     &  Encoder     &  Decoder         & Encoder       & Decoder          & mIoU  & Rice  & Corn  & Soybean & Others \\
        \hline   
        baseline     & ConvNeXt-B   & UperNet          & —              & —               & 77.89 & 86.47 & 78.36 & 68.47   & 78.26 \\
        TransLU      & ConvNeXt-B   & UperNet+BHCCM    & ConvNeXt-B     & UperNet+BHCCM   & 80.58 & 88.60 & 81.00 & 70.80   & 81.90 \\
        \hline    
        baseline     & DeiT3-B      & UperNet          & —              & —               & 75.51 & 85.70 & 74.98 & 65.14   & 76.22 \\
        TransLU      & DeiT3-B      & UperNet+BHCCM    & DeiT3-B        & UperNet+BHCCM   & 78.17 & 87.91 & 77.38 & 67.07   & 80.33\\
        \hline    
        baseline     & MSCA-B       & SegNeXt          & —              & —               & 76.28 & 85.47 & 75.96 & 66.56   & 77.14 \\
        TransLU      & MSCA-B       & SegNeXt+BHCCM    & MSCA-B         & SegNeXt+BHCCM   & 78.92 & 88.08 & 78.81 & 68.41   & 80.37 \\
        \bottomrule
    \end{tabular}
    }
\end{table*}

\subsubsection{Qualitative Results}
\begin{figure*}[!tb]
	\begin{center}
		\centerline{\includegraphics[width=0.90\textwidth]{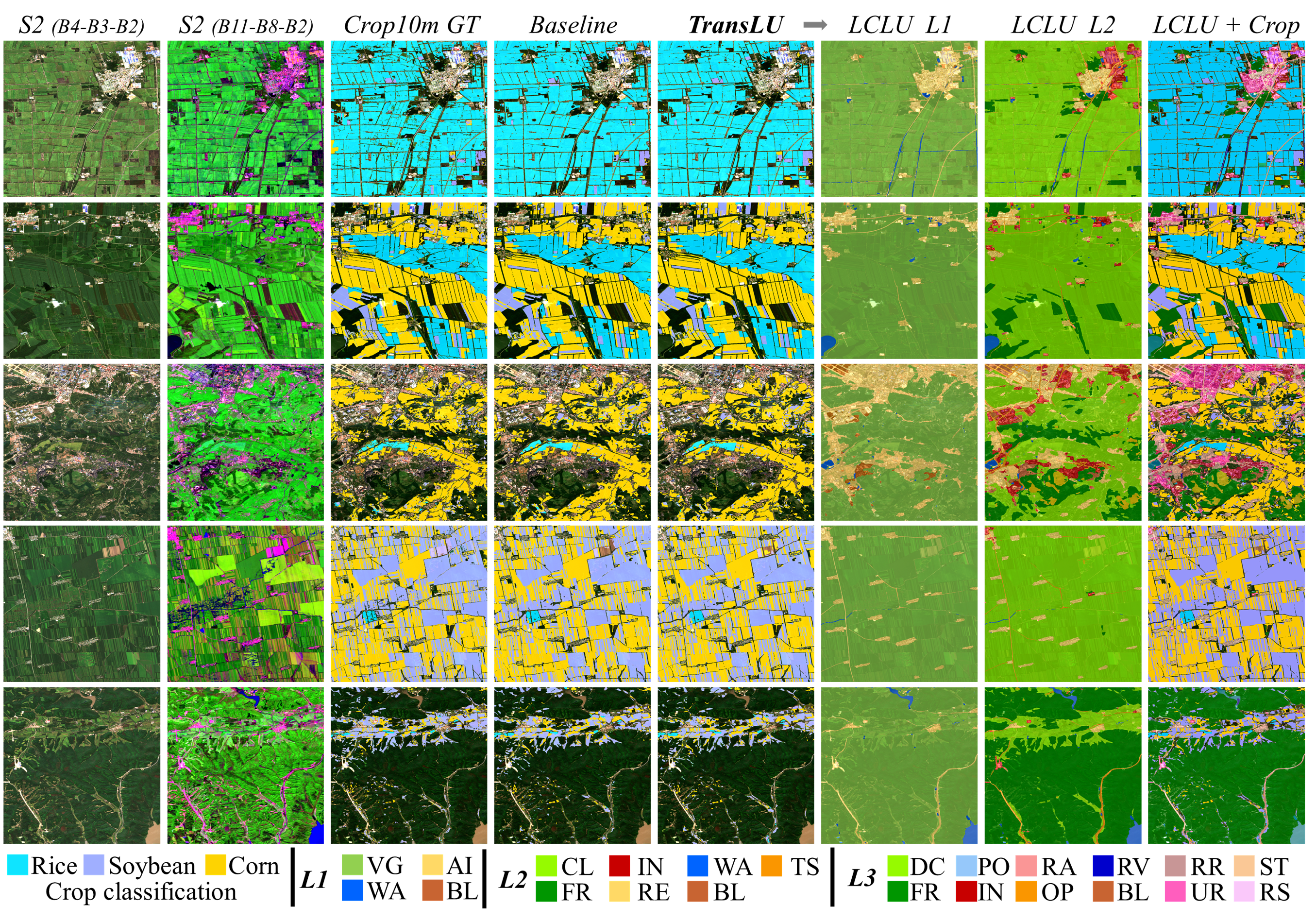}}
        \caption{Visualization of crop classification results on the Crop10m dataset. This figure illustrates the transfer of the hierarchical LCLU classification model (ConvNeXt-B + BHCCM) to the crop classification task. The first two columns display Sentinel-2 images with different band combinations, highlighting the importance of spectral features for crop identification. \textit{Baseline} refers to the flat ConvNeXt-B model trained from scratch, while \textit{TransLU} denotes the results obtained using the proposed dual-branch transfer framework. \textit{LCLU L1} and \textit{LCLU L2} represent the hierarchical land use predictions inferred from Branch 2 of TransLU. The last column displays the comprehensive full-category results, encompassing both crop classification and fine-grained LCLU details, jointly output by the two branches of TransLU.}
    \label{crop10m_result}
	\end{center}
\end{figure*}

Beyond facilitating model transfer, TransLU demonstrates robust potential for dynamic category expansion. It reformulates crop classification as a fine-grained extension of the LCLU hierarchy, achieved by dynamically broadening the tree-structured taxonomy to meet specific application requirements. As depicted in Fig.~\ref{fig-overall_arch}(d), specific crops are modeled as downstream sub-nodes of cropland. This hierarchical refinement capability is extensible to diverse scenarios, such as refining \textit{``Artificial Surfaces $\rightarrow$ Industrial Area $\rightarrow$ Storage Tanks''} or \textit{``Artificial Surfaces $\rightarrow$ Transportation $\rightarrow$ Roads $\rightarrow$ Highway''}. Qualitatively, Fig.~\ref{crop10m_result} validates this approach, visualizing the precise subdivision of cropland into rice, corn, and soybeans as a consistent L3 extension.

\subsection{Transfer LCLU Models Across Heterogeneous Hierarchy}
Table~\ref{WHDLD-results-table} presents the results of transferring a pretrained hierarchical LCLU model to a heterogeneous classification system using the proposed TransLU on the WHDLD dataset. In this experiment, the hierarchical LCLU model in Branch~2 was trained on the Google Earth imagery subset from the MM-5B dataset. The experimental results indicate that applying the TransLU framework to transfer the hierarchical LCLU model to a task with a different label hierarchy significantly improves performance compared to the baseline.

\begin{table}[tb]
    \centering
    \caption{Performance comparison of the proposed TransLU on the WHDLD dataset.}
    \label{WHDLD-results-table}
    \renewcommand\arraystretch{1.1} 
    \fontsize{7}{8}\selectfont      
    \begin{adjustbox}{width=0.85\linewidth}
    \begin{tabular}{l|l|l|c}
        \toprule
        \textbf{Method} & \textbf{Encoder}                                       & \textbf{Decoder}   & \textbf{mIoU}     \\
        \midrule     
        baseline        & \multirow{2}{*}{ConvNeXt-B\cite{convnext_2022_cvpr}}   &  UperNet           & 64.95             \\
        TransLU         &                                                        &  UperNet+BHCCM     & 68.36             \\
        \hline                   
        baseline        & \multirow{2}{*}{DeiT3-B\cite{deit3_2022_ECCV}}         &  UperNet           & 64.12             \\
        TransLU         &                                                        &  UperNet+BHCCM     & 66.77             \\
        \hline                 
        baseline        & \multirow{2}{*}{SegNeXt-B\cite{segnext_2022_NeurIPS}}  &  SegNeXt           & 65.34             \\
        TransLU         &                                                        &  SegNeXt+BHCCM     & 67.49             \\
        \bottomrule 
    \end{tabular}
    \end{adjustbox}
\end{table}

Specifically, ConvNeXt-B achieves the most notable gains, with mIoU increasing by 3.41\% (from 64.95\% to 68.36\%). Similarly, the Transformer-based DeiT3-B exhibits a substantial improvement of 2.65\%, reaching 66.77\%. Furthermore, the lightweight SegNeXt-B also benefits from the proposed method, achieving a 2.15\% gain in mIoU. These consistent improvements across diverse backbones demonstrate that TransLU is highly adaptable to different network architectures. Moreover, the results confirm that transferring a well-trained hierarchical LCLU model via TransLU is superior to training from scratch, offering a highly effective paradigm for cross-domain LCLU applications.

\begin{figure}[tb]
    \centering 
    \includegraphics[width=0.90\linewidth]{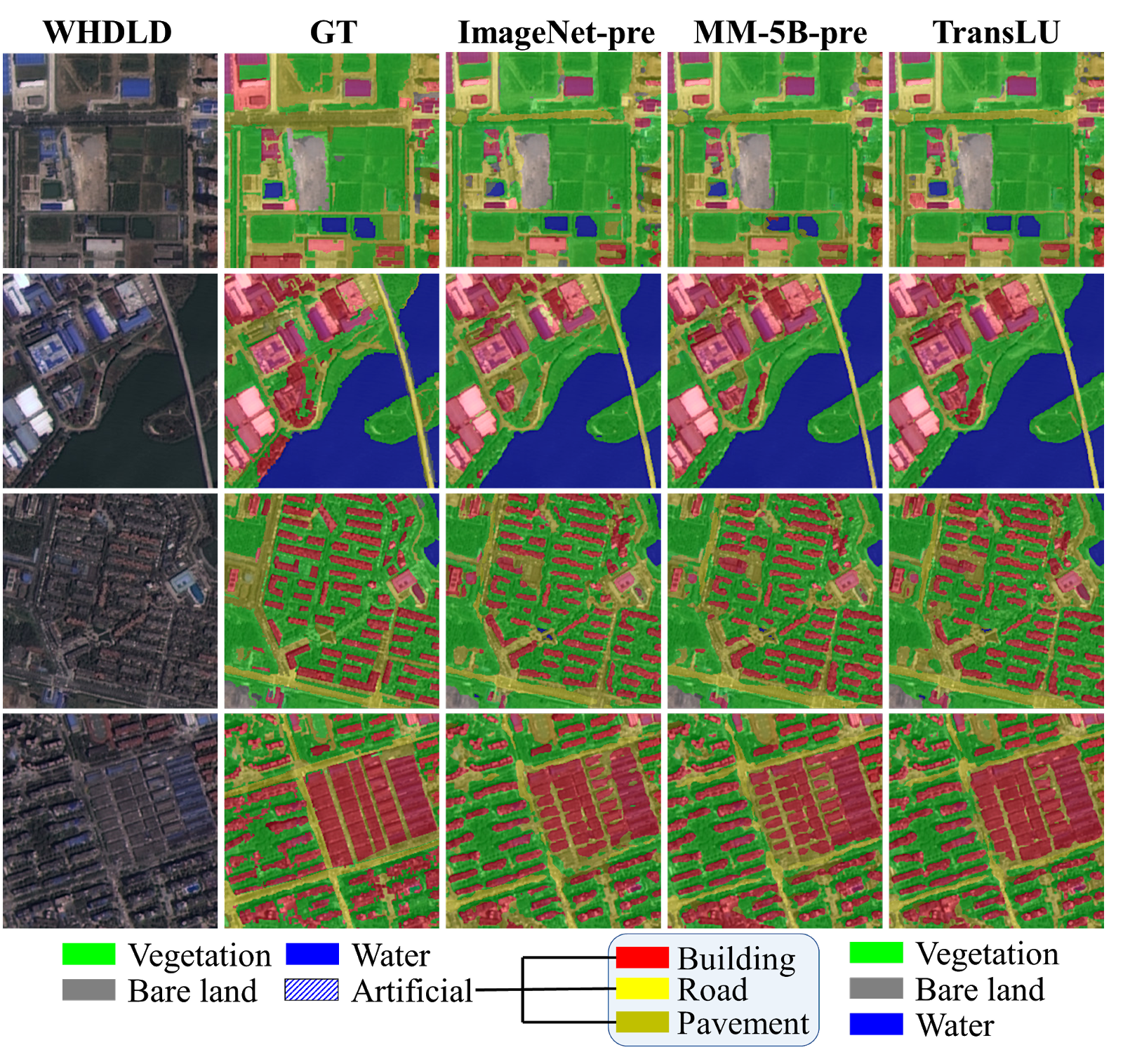}
    \caption{Qualitative comparison on the WHDLD dataset using ConvNeXt-B. The proposed TransLU demonstrates superior segmentation details compared to baselines initialized with ImageNet and MM-5B weights. The legend illustrates the tree-structured hierarchy used in TransLU.}
    \label{WHDLD-result}
\end{figure}

Collectively, these findings demonstrate the high adaptability of TransLU across diverse network architectures and confirm that transferring a pretrained hierarchical LCLU model yields superior performance compared to standard training baselines. This establishes a promising paradigm for efficiently adapting well-trained hierarchical models to heterogeneous domains. Furthermore, the qualitative visualization in Fig.~\ref{WHDLD-result}, comparing ConvNeXt-B under the baseline and TransLU frameworks, intuitively highlights the method's effectiveness in refining classification details.

\subsection{Ablation Study of BHCCM and the loss $\mathcal{L}_\text{HSC}$}
\label{SEC:ablation-BHCCM}
In this ablation study, we evaluate the contribution of the bidirectional semantic consistency constraints built into BHCCM and the effectiveness of the proposed loss $\mathcal{L}_\text{HSC}$. As reported in Table~\ref{ablation-BHCCM-table}, we adopt ConvNeXt-Base with a flat UperNet as the baseline. On the GaoFen-2 source of the MM-5B dataset, the baseline achieves mIoU scores of 73.39\%, 82.38\%, and 93.73\% at $L3$, $L2$, and $L1$, respectively. 
To systematically isolate the contributions of architectural designs and loss constraints, we construct eight comparative settings, as detailed below:

\begin{table}[tb]
    \centering
    \caption{mIoU Performance Ablation Analysis of BHCCM and $\mathcal{L}_\text{HSC}$ on the GaoFen-2 Source of MM-5B Dataset.}
    \label{ablation-BHCCM-table}
    \fontsize{8}{9}\selectfont 
    \renewcommand\arraystretch{1.1} 
    \begin{adjustbox}{width=0.95\linewidth}
    \begin{tabular}{c|l|c|c c c}
        \toprule
        \textbf{ID}  & \textbf{Method}            & \textbf{Loss}              & $\bm{L3}$ & $\bm{L2}$ & $\bm{L1}$ \\
        \midrule
            0        & Flat Baseline              & $\mathcal{L}_\text{CE}$    & 73.39  & 82.38  & 93.73    \\ 
            1        & + BHCCM (without fusion)   & $\mathcal{L}_\text{HCE}$   & 73.04  & 82.49  & 93.81    \\ 
            2        & + BHCCM (coarse-to-fine)   & $\mathcal{L}_\text{HCE}$   & 73.23  & 82.56  & 93.85    \\ 
            3        & + BHCCM (fine-to-coarse)   & $\mathcal{L}_\text{HCE}$   & 73.17  & 82.58  & 93.94    \\ 
            4        & + BHCCM (bidirectional)    & $\mathcal{L}_\text{HCE}$   & 73.26  & 82.64  & 93.93    \\ 
            \hline
            5        & + BHCCM (without fusion)   & $\mathcal{L}_\text{HSC}$   & 73.47  & 82.67  & 94.00    \\ 
            6        & + BHCCM (coarse-to-fine)   & $\mathcal{L}_\text{HSC}$   & 73.94  & 82.74  & 94.02    \\ 
            7        & + BHCCM (fine-to-coarse)   & $\mathcal{L}_\text{HSC}$   & 73.66  & 82.80  & 94.17    \\ 
            8        & + BHCCM (bidirectional)    & $\mathcal{L}_\text{HSC}$   & 74.13  & 82.79  & 94.21    \\ 
    \bottomrule
    \end{tabular}
    \end{adjustbox}
\end{table}

\begin{enumerate}[label=(\arabic*), leftmargin=*]
    \item \textbf{Without fusion ($\mathcal{L}_{\text{HCE}}$):} Utilizes three independent 2D convolution heads. The mIoU at $L3$ is 0.35\% lower than the baseline, while $L2$ and $L1$ are 0.11\% and 0.08\% higher, respectively.
    
    \item \textbf{Coarse-to-fine ($\mathcal{L}_{\text{HCE}}$):} Involves only coarse-to-fine information flow. The mIoU at $L3$ is 0.16\% lower than the baseline, whereas $L2$ and $L1$ surpass the baseline by 0.18\% and 0.12\%.
    
    \item \textbf{Fine-to-coarse ($\mathcal{L}_{\text{HCE}}$):} Involves only fine-to-coarse information flow. The mIoU at $L3$ is 0.22\% lower than the baseline, while $L2$ and $L1$ exceed the baseline by 0.20\% and 0.21\%.
    
    \item \textbf{Bidirectional ($\mathcal{L}_{\text{HCE}}$):} Contains complete bidirectional information flow. The mIoU at $L3$ remains 0.13\% lower than the baseline, but $L2$ and $L1$ are 0.26\% and 0.20\% higher.
    
    \item \textbf{Without fusion ($\mathcal{L}_{\text{HSC}}$):} Uses independent heads constrained by the proposed $\mathcal{L}_{\text{HSC}}$. The mIoU values at $L3$, $L2$, and $L1$ are all higher than the baseline, improving by 0.08\%, 0.29\%, and 0.27\%, respectively.
    
    \item \textbf{Coarse-to-fine ($\mathcal{L}_{\text{HSC}}$):} Incorporates coarse-to-fine flow with HSC loss. The mIoU values at $L3$, $L2$, and $L1$ surpass the baseline by 0.55\%, 0.36\%, and 0.29\%.
    
    \item \textbf{Fine-to-coarse ($\mathcal{L}_{\text{HSC}}$):} Incorporates fine-to-coarse flow with HSC loss. The mIoU values at $L3$, $L2$, and $L1$ exceed the baseline by 0.27\%, 0.42\%, and 0.44\%.
    
    \item \textbf{Bidirectional ($\mathcal{L}_{\text{HSC}}$):} Integrates complete bidirectional information flow with HSC loss. This setting achieves the optimal performance, with mIoU values at $L3$, $L2$, and $L1$ surpassing the baseline by 0.74\%, 0.41\%, and 0.48\%, respectively.
\end{enumerate}

Based on the results reported in Table~\ref{ablation-BHCCM-table}, we conduct a detailed analysis to investigate the impact of different configurations step-by-step.

Comparing Exp~1 with the Baseline reveals the impact of independent predictions. We observe that utilizing independent convolution heads significantly improves $L1$ and $L2$ performance but leads to a noticeable drop in $L3$. This indicates that explicitly learning coarse-grained features is superior to the baseline's approach of simply merging fine-grained predictions. However, the drop in $L3$ suggests that under the standard $\mathcal{L}_{\text{HCE}}$, a multi-task optimization trade-off occurs, where the model's capacity is diverted to optimize additional coarse levels, slightly compromising the fine-grained prediction.

Introducing directional information flows (Exp~2--4) partially mitigates this trade-off. Specifically, the coarse-to-fine strategy (Exp~2) explicitly transfers the strong discriminative semantics from parent classes to finer levels, which leads to improved accuracy at $L2$ and $L3$. In contrast, the fine-to-coarse strategy (Exp~3) feeds rich detailed features back to coarser levels, effectively enhancing performance at $L1$ and $L2$. Furthermore, the Bidirectional strategy (Exp~4) combines both benefits, yielding better overall results than unidirectional flows. Nevertheless, under $\mathcal{L}_{\text{HCE}}$, the $L3$ performance in all these settings remains below the baseline, indicating that feature fusion alone is insufficient to fully resolve the task conflict.

A crucial turning point is observed when replacing $\mathcal{L}_{\text{HCE}}$ with the proposed $\mathcal{L}_{\text{HSC}}$. In Exp~5 (without fusion), $L3$ performance surpasses the baseline for the first time. This demonstrates that $\mathcal{L}_{\text{HSC}}$ effectively enforces hierarchical semantic consistency, transforming the competing multi-level tasks into a synergistic optimization process. By learning valid semantic path distributions, the model can simultaneously optimize $L1$, $L2$, and $L3$ without degradation.

Finally, the integration of distinct information flows with $\mathcal{L}_{\text{HSC}}$ yields consistent performance gains. Similar to the trends in $\mathcal{L}_{\text{HCE}}$ settings, the coarse-to-fine flow primarily enhances $L3$ and $L2$, whereas the fine-to-coarse flow predominantly favors $L2$ and $L1$. Most notably, the full bidirectional BHCCM coupled with $\mathcal{L}_{\text{HSC}}$ (Exp~8) attains optimal performance across all hierarchical levels. This validates that the dual-way feature enhancement and the hierarchical consistency constraint are mutually reinforcing. It is this joint deployment that unlocks the full potential of the proposed BHCCM, resulting in substantial improvements over the baseline.

These ablation studies reveal a strong coupling between the proposed BHCCM structure and the hierarchical semantic consistency loss $\mathcal{L}_{\text{HSC}}$. First, when constrained only by the standard cross-entropy loss $\mathcal{L}_{\text{HCE}}$, simply introducing feature interaction fails to yield positive gains. As shown in Exp~2 and Exp~3, employing unidirectional fusion results in suboptimal performance at the finest $L3$ level (lagging behind the baseline by 0.16\%--0.22\%). We attribute this to the optimization conflict: extending the model to output three distinct hierarchical levels requires the backbone to simultaneously satisfy divergent semantic granularities. When supervised solely by independent $\mathcal{L}_{\text{HCE}}$, these inconsistent objectives pull feature extraction in conflicting directions. Consequently, even with full bidirectional interaction Exp~4, the $L3$ mIoU (73.26\%) remains inferior to the flat baseline (73.39\%), indicating that feature fusion alone cannot resolve this semantic conflict.

Conversely, the efficacy of BHCCM is fully unlocked only when paired with $\mathcal{L}_{\text{HSC}}$. As evidenced by the comparing between Exp~1 ($\mathcal{L}_{\text{HCE}}$) and Exp~5 ($\mathcal{L}_{\text{HSC}}$), substituting the loss function triggers a decisive performance reversal---shifting from a 0.35\% degradation to a 0.08\% improvement over the baseline. This demonstrates that $\mathcal{L}_{\text{HSC}}$ is indispensable for aligning the optimization objectives across hierarchies. Furthermore, under $\mathcal{L}_{\text{HSC}}$ supervision, the bidirectional design Exp~8 achieves the most significant gain of 0.74\% at $L3$. This confirms that the synergy between the bidirectional architecture and the consistency loss is essential for maximizing representational capability. Note: To strictly evaluate the model's representational capability, all ablation results reported above are computed without the JSPS inference strategy.

\subsection{Complexity Analysis of BHCCM}

Furthermore, to assess the overhead incurred by the multi-level prediction mechanism and the internal Merging Blocks, we analyzed the computational consumption of BHCCM. Table~\ref{tab:BHCCM_complexity_analysis} details the increase in parameters and FLOPs on ConvNeXt and SegNeXt architectures, utilizing an input resolution of $640 \times 640 \times 4$. The results indicate that BHCCM introduces only a marginal increase in computational cost, maintaining the model's efficiency while significantly boosting performance.

Considering the substantial performance gains and the capability for strictly consistent multi-level inference, we argue that this negligible computational overhead is highly cost-effective for practical deployment. The consistent efficacy observed across varying backbone sizes further validates the scalability of our approach. Consequently, for latency-sensitive scenarios, employing a lightweight backbone equipped with BHCCM offers an optimal strategy, achieving high-speed inference without compromising the robustness of hierarchical interpretation.

\begin{table}[t]
    \centering
    \caption{Computational Complexity Analysis on Parameters and FLOPs. The input shape is set to $640 \times 640 \times 4$ for all models.}
    \label{tab:BHCCM_complexity_analysis}
    \renewcommand{\arraystretch}{1.2} 
    \resizebox{0.95\linewidth}{!}{ 
        \begin{tabular}{c|l|l l|c}
            \toprule
            \textbf{Backbone} & \textbf{Decoder} & \textbf{Parameters} & \textbf{FLOPs} &  \textbf{mIoU}\\
            \midrule
            \multirow{2}{*}{MSCA-S}     & SegNeXt          & 13.8993M  & 24.3674G     & 74.29\\
                                        & SegNeXt + BHCCM  & 13.9043M  & 24.3963G     & 74.77\\
            \cline{1-5}
             \multirow{2}{*}{MSCA-B}    & SegNeXt          & 27.5689M  & 50.8267G     & 74.74\\
                                        & SegNeXt + BHCCM  & 27.5772M  & 50.8769G     & 75.72\\
            \cline{1-5}
            \multirow{2}{*}{MSCA-L}     & SegNeXt          & 48.7915M  & 102.4830G    & 76.23\\
                                        & SegNeXt + BHCCM  & 48.8065M  & 102.5758G    & 76.59\\
            \midrule
            \multirow{2}{*}{ConvNeXt-B} & UperNet          & 120.8186M & 455.7063G    & 73.39\\
                                        & UperNet + BHCCM  & 120.8269M & 455.9071G    & 74.16\\
            \cline{1-5}
            \multirow{2}{*}{ConvNeXt-L} & UperNet          & 233.1216M & 613.4438G    & 75.32\\
                                        & UperNet + BHCCM  & 233.1300M & 613.6447G    & 76.18\\
            \bottomrule
        \end{tabular}
    }
\end{table}

In conclusion, these results clearly demonstrate that the bidirectional semantic consistency constraints and the hierarchical semantic consistency loss in BHCCM significantly enhance the performance of hierarchical LCLU classification. Crucially, these substantial gains are achieved with only a marginal increase in parameters, underscoring the lightweight and efficient design of BHCCM.

\begin{table}[b]
    \centering
    \caption{Ablation Study of the JSPS Strategy on the GaoFen-2 Source of MM-5B Dataset in terms of mIoU.}
    \label{tab:ablation-JSPS}
    \renewcommand{\arraystretch}{1.1} 
    \fontsize{7}{8}\selectfont 
    \resizebox{0.85\linewidth}{!}{ 
        \begin{tabular}{l|c|c c c}
            \toprule
            \textbf{Model}                 & \textbf{JSPS} & $\bm{L3}$ & $\bm{L2}$ & $\bm{L1}$ \\
            \midrule
            \multirow{2}{*}{SegNeXt-B}     &               & 75.68       & 83.48       & 94.56 \\
                                           & \cmark        & 75.72       & 83.63       & 94.62 \\
            \hline            
            \multirow{2}{*}{ConvNeXt-B}    &               & 74.13       & 82.79       & 94.21 \\
                                           & \cmark        & 74.16       & 82.84       & 94.22 \\
            \hline            
            \multirow{2}{*}{DeiT3-B}       &               & 73.82       & 82.38       & 94.39 \\
                                           & \cmark        & 73.87       & 82.49       & 94.41 \\
            \bottomrule
        \end{tabular}
    }
\end{table}

\subsection{Ablation Study of JSPS Strategy}
We validate the effectiveness of the proposed Joint Score-based Path Selection (JSPS) strategy using the GaoFen-2 subset of the MM-5B dataset. As detailed in Table~\ref{tab:ablation-JSPS}, we benchmark the strategy across three different backbones: SegNeXt-B, ConvNeXt-B, and DeiT3-B. The results demonstrate that applying JSPS as a post-processing step consistently improves performance, yielding mIoU gains ranging from 0.02\% to 0.15\% across all three hierarchical levels. Although these numerical improvements are modest, which suggests that our base BHCCM already achieves strong hierarchical consistency, the value of JSPS extends beyond quantitative metrics. Crucially, it ensures strict adherence to the predefined tree-structured hierarchy, guaranteeing semantically valid and logically consistent predictions for hierarchical LCLU classification tasks.

\begin{table}[tb]
    \centering
    \caption{Ablation analysis of CDKS and CDSA on the Crop10m dataset using ConvNeXt-B.}
    \label{ablation-CDKS-CDSA}
    \renewcommand{\arraystretch}{1.25} 
    \fontsize{7}{9}\selectfont 
    \begin{adjustbox}{width=0.90\linewidth}
    \begin{tabular}{c|c c l}
        \toprule
        \textbf{Branch1 Init.}   & \textbf{CDKS}    & \textbf{CDSA}  &\textbf{mIoU (\%)} \\  
        \midrule  
        ImageNet-21k             &                  &                &77.89   \\
        \hline 
        \multirow{4}{*}{\makecell{MM-5B \\ (Sentinel-2)}}   &   &    &78.63 (+0.74\%) \\
                                & \cmark           &                 &79.06 (+1.17\%) \\
                                &                  & \cmark          &79.27 (+1.38\%) \\
                                & \cmark           & \cmark          &80.58 (+2.69\%) \\

    \bottomrule
    \end{tabular}
    \end{adjustbox}
\end{table}

\subsection{Ablation Study of CDKS and CDSA}

The proposed TransLU framework consists of two key components: the Cross-Domain Knowledge Sharing (CDKS) and the Cross-Domain Semantic Alignment (CDSA). To evaluate their effectiveness, we conducted ablation studies on the Crop10m crop classification task using the ConvNeXt-B backbone. The results are summarized in Table~\ref{ablation-CDKS-CDSA}.

First, we investigate the impact of weight initialization. As shown in the table, replacing the generic ImageNet-21k weights with the LCLU model weights pre-trained on the Sentinel-2 subset of the MM-5B dataset brings a clear performance gain. Even without any specific transfer modules, the mIoU improves from 77.89\% to 78.63\%, a gain of 0.74\%. This confirms that the MM-5B pre-training provides superior feature initialization adapted to the spectral characteristics of Sentinel-2 imagery.

Next, we evaluate the individual contributions of CDKS and CDSA based on the MM-5B initialization. When integrating CDKS into the framework, the model achieves an mIoU of 79.06\%, representing a 1.17\% improvement over the baseline. This indicates that distilling knowledge from the fixed Branch~2 effectively guides the learning of Branch~1. Similarly, applying CDSA alone yields an mIoU of 79.27\%, surpassing the baseline by 1.38\%. This result highlights the importance of semantic alignment in facilitating the transfer from general LCLU to specific crop categories.

Finally, the best performance is achieved when both components are jointly employed. The full TransLU framework reaches an mIoU of 80.58\%, a substantial increase of 2.69\% compared to the baseline. This significant boost demonstrates that CDKS and CDSA are mutually reinforcing: CDKS provides robust feature guidance, while CDSA ensures semantic consistency, collectively facilitating the efficient transfer of the hierarchical LCLU model to the downstream crop classification task.

\section{Conclusion}
\label{section:Conclusion}
In this paper, we presented HieraRS, a novel hierarchical interpretation paradigm designed to address critical challenges in multi-granularity LCLU classification and cross-domain task transfer. To achieve semantically consistent hierarchical classification, we proposed the Bidirectional Hierarchical Consistency Classification Model (BHCCM). By embedding inherent tree-structured class relationships and enforcing the proposed Hierarchical Semantic Consistency loss ($\mathcal{L}_{\text{HSC}}$), BHCCM effectively transforms conventional flat models into hierarchical predictors. This integration ensures prediction consistency across levels while simultaneously improving accuracy at each granularity. To facilitate model transferability to cross-domain EO tasks, we introduced TransLU, a dual-branch framework equipped with Cross-Domain Knowledge Sharing (CDKS) and Cross-Domain Semantic Alignment (CDSA). TransLU enables the efficient adaptation of LCLU models to heterogeneous hierarchical classification systems, significantly promoting generalization and reducing reliance on labeled data in target domains. Furthermore, we contributed MM-5B to the remote sensing community, a large-scale multi-source and multi-resolution hierarchical LCLU dataset with pixel-level annotations. Comprising imagery from three distinct sensors, MM-5B covers diverse spatial resolutions and spectral characteristics. Extensive experiments on MM-5B, Crop10m, and WHDLD demonstrate that the HieraRS paradigm delivers outstanding performance in both hierarchical LCLU classification and cross-domain transfer tasks.

In future work, we plan to enrich the MM-5B dataset with additional modalities, such as Hyperspectral Imagery (HSI) and Synthetic Aperture Radar (SAR). This expansion will allow us to further investigate the adaptability of HieraRS to multi-modal approaches. Additionally, we aim to extend the scope of HieraRS from single-temporal segmentation to multi-temporal analysis tasks, such as Change Detection, which require processing time-series imagery. We believe that these extensions will further solidify HieraRS as a versatile and reliable framework for diverse remote sensing interpretation tasks.

\bibliographystyle{IEEEtran}
\bibliography{HieraRS_ref}
\vfill

\end{document}